\pdfoutput=1
\documentclass[11pt]{article}
\usepackage[table]{xcolor}
\usepackage[final]{acl}
\usepackage{times}
\usepackage{latexsym}
\usepackage[T1]{fontenc}
\usepackage{longtable}
\usepackage[utf8]{inputenc}
\usepackage{microtype}
\usepackage{inconsolata}
\usepackage{algorithm}
\usepackage{bm}
\usepackage{booktabs}
\usepackage{makecell}
\usepackage[table]{xcolor}

\usepackage{wrapfig}
\usepackage[utf8]{inputenc}
\usepackage{booktabs}
\usepackage[most]{tcolorbox}
\usepackage{verbatim}
\usepackage{makecell}
\usepackage{multirow}
\usepackage{tabularx}
\usepackage{subcaption}
\usepackage{colortbl}
\usepackage{algorithmicx}
\usepackage{float}
\usepackage{stfloats}  
\usepackage{booktabs}
\usepackage{makecell}
\usepackage{amssymb}
\usepackage{multicol}
\usepackage{graphicx}
\usepackage{array}
\usepackage{enumitem}
\usepackage{svg}
\usepackage{pdfpages}
\usepackage{natbib}
\usepackage{hyperref}
\usepackage{url}
\usepackage{amsmath}
\usepackage{amssymb}
\usepackage{lipsum}
\usepackage{caption}
\usepackage{fontawesome5}  

\usepackage{xcolor}
\usepackage{pifont}

\newcommand{\cmark}{\textcolor{green!60!black}{\ding{51}}} 
\newcommand{\xmark}{\textcolor{red}{\ding{55}}}           

\newcommand{\sref}[1]{\S\ref{#1}}

\title{SurveyGen: Quality-Aware Scientific Survey Generation with Large Language Models}

\author{
  Tong Bao\textsuperscript{1,2}\thanks{This work was partly done at the University of Alberta. \\ \hspace*{1.5em}\dag Corresponding authors},
  Mir Tafseer Nayeem\textsuperscript{2}, 
  Davood Rafiei\textsuperscript{2 \dag}, 
  Chengzhi Zhang\textsuperscript{1 \dag}  \\ %
  \textsuperscript{1}Nanjing University of Science and Technology\\
  \textsuperscript{2}University of Alberta\\
  \texttt{\{tbao,zhangcz\}@njust.edu.cn}, \texttt{\{mnayeem,drafiei\}@ualberta.ca}
}

\begin{document}
\maketitle
\begin{abstract}
Automatic survey generation has emerged as a key task in scientific document processing. While large language models (LLMs) have shown promise in generating survey texts, the lack of standardized evaluation datasets critically hampers rigorous assessment of their performance against human-written surveys. In this work, we present \texttt{\textbf{SurveyGen}}, a large-scale dataset comprising over 4,200 human-written surveys across diverse scientific domains, along with 242,143 cited references and extensive quality-related metadata for both the surveys and the cited papers. Leveraging this resource, we build QUAL-SG, a novel quality-aware framework for survey generation that enhances the standard Retrieval-Augmented Generation (RAG) pipeline by incorporating quality-aware indicators into literature retrieval to assess and select higher-quality source papers. Using this dataset and framework, we systematically evaluate state-of-the-art LLMs under varying levels of human involvement—from fully automatic generation to human-guided writing. Experimental results and human evaluations show that while semi-automatic pipelines can achieve partially competitive outcomes, fully automatic survey generation still suffers from low citation quality and limited critical analysis\footnote{Code and data are available at \url{https://github.com/tongbao96/SurveyGen}}.

\end{abstract}

\section{Introduction}
Survey articles play a crucial role in summarizing previous research on a specific topic, providing a comprehensive understanding of the field, and supporting further advancements \cite{Torraco2005}. However, writing a survey is a complex task as it typically requires summarizing hundreds of relevant studies. The rapid expansion of academic publications further adds to the difficulty, making it increasingly challenging for researchers to keep up with the latest findings. Given these challenges, the development of automatic survey generation systems has become a key focus in the field of scientific document processing \cite{wang2024autosurvey}.

\begin{figure}[t] 
  \centering 
\includegraphics[width=\columnwidth, clip]{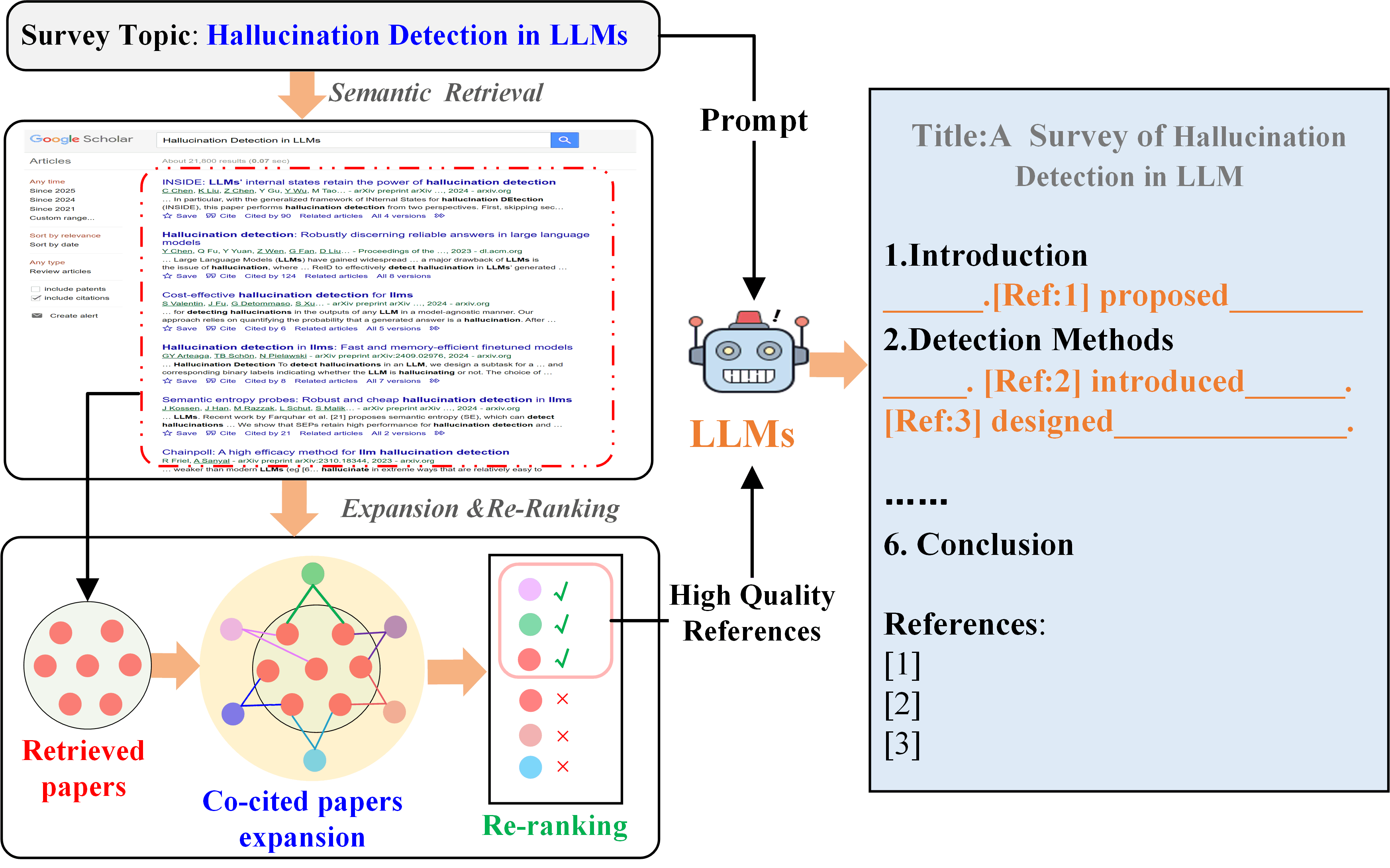} 
  \caption{Overview of the proposed QUAL-SG: a quality-aware framework that leverages semantic retrieval and citation expansion to select high-quality literature and support more reliable survey generation.}
  \label{fig:figure1} 
\end{figure}

Leveraging the strong text generation capabilities of large language models (LLMs) \cite{brown2020language}, recent studies on automatic survey generation have adopted Retrieval-Augmented Generation (RAG) techniques~\cite{izacard2022few, borgeaud2022improving, gao2023retrieval, agarwal2024litllm} to augment them with external knowledge sources, yielding promising results \cite{wang2024autosurvey, tang2024llms, wu2024automated, liang2025surveyx}. However, these approaches fall short in two key aspects: \textbf{(1)} the retrieval of high-quality literature, and \textbf{(2)} the rigorous evaluation against the human-authored gold standard.

In the retrieval stage, most prior works rely on semantic and syntactic similarity between user-provided survey topics and publication abstracts to identify relevant studies. These methods typically do not assess the quality, impact, or influence of the retrieved literature. Yet, a well-crafted survey is expected to not only summarize existing research but also highlight seminal works and major advancements in the field \cite{snyder2019literature,paul2020art,kanellos2019impact}. As a result, retrieving articles based purely on textual relevance risks including low-impact or marginal studies, which in turn diminishes the quality and credibility of the generated survey.

In the evaluation stage, although recent works have employed both automatic and human evaluations to assess LLM-generated surveys \cite{wu2024automated, lai2024instruct, agarwal2024llms, wang2024autosurvey, liang2025surveyx}, the lack of large-scale benchmarks has hindered systematic comparisons with human-written surveys, which remain the gold standard. In particular, critical evaluation dimensions such as citation quality, structural consistency, and domain-specific variation remain underexplored. Without comprehensive benchmarks, it is difficult to rigorously evaluate whether LLM-generated surveys meet the quality, reliability, and scholarly standards expected in academic writing.

To address the above limitations, we first introduce \texttt{\textbf{SurveyGen}}, a large-scale dataset comprising over 4,200 human-written surveys from the Semantic Scholar Open Research Corpus (S2ORC; \citeauthor{lo2020s2orc}, \citeyear{lo2020s2orc}), along with 242,143 cited references within these surveys and extensive metadata for all referenced papers for evaluation purposes. Building on this resource, we propose QUAL-SG, a novel quality-aware literature retrieval framework designed to enhance the reliability and relevance of retrieved articles for survey generation. As shown in Figure \ref{fig:figure1}, QUAL-SG first expands the candidate reference pool via citation graph analysis, then re-ranks articles based on quality indicators, ensuring both citation reliability and broad literature coverage for survey generation. We design three targeted tasks, each equipped with domain-appropriate evaluation metrics, to provide a comprehensive analysis of LLMs' effectiveness across different stages of the survey generation pipeline. 

Our contributions can be summarized as follows:

\noindent\hspace*{1em}{\Large\textbullet}\;\  We introduce \texttt{\textbf{SurveyGen}}, a large-scale dataset comprising over 4,200 human-written surveys with section-level structures, including cited references and rich metadata capturing citation performance, author influence, and venue reputation. \texttt{\textbf{SurveyGen}} supports comprehensive evaluation across content quality, citation quality, and structural consistency in scientific survey generation tasks.

\noindent\hspace*{1em}{\Large\textbullet}\;\ We propose \textbf{QUAL-SG}, a novel quality-aware framework that extends Naive-RAG by incorporating literature quality assessment into the survey generation pipeline. Our results show that QUAL-SG significantly improves citation reliability and enhances the overall content quality and structure consistency of the generated surveys.

\noindent\hspace*{1em}{\Large\textbullet}\;\ We benchmark several state-of-the-art LLMs under varying levels of involvement in the survey generation process and conduct extensive evaluations—both automatic and human—to analyze model performance, identify key limitations, and offer actionable insights for future research on LLM-assisted academic writing.

\section{Approach}
\label{sec:approach}
In this section, we first introduce the design of the survey generation tasks (\sref{sec:task-design}), then present our \texttt{\textbf{SurveyGen}} dataset (\sref{sec:dataset-construction},  \sref{sec:Indicators-Supplement}) and the proposed QUAL-SG framework (\sref{sec:QUAL-SG}).

\subsection{Task Design}  
\label{sec:task-design}
Given that humans may engage LLMs at different stages during survey generation depending on their specific goals  (e.g., literature retrieval, outline generation, or content drafting), the level of involvement can vary considerably. We define \emph{three} representative tasks to systematically evaluate LLMs' generation capabilities across these different levels: \textbf{(1) Fully LLM-based}, \textbf{(2) RAG-based}, and \textbf{(3) Human-guided Survey Generation}. 

The \underline{\textit{distinct focus}} of the three tasks is as follows: Task 1 evaluates LLM's capability to generate a complete survey without access to external sources; Task 2 evaluates its performance under the standard RAG setting, where relevant literature is first retrieved from an external database and then used to support survey generation; and Task 3 evaluates the generated survey when LLMs are provided with human-selected references and a human-written predefined outline, simulating a fully guided writing setting. The definitions of these three tasks are detailed below:

\textbf{Task 1: Fully LLM-based Survey Generation:}  \label{task1}
Given only a survey topic \( t_i \), the LLMs are prompted to generate the entire survey, including a structured outline, corresponding content, and a relevant list of references. No external documents or human-crafted materials are provided. 

\textbf{Task 2: RAG-based Survey Generation:}  \label{task2} 
This task follows the standard RAG pipeline, where a retriever identifies relevant literature from an external database, and a generator writes the survey's outline and content. Given a survey topic \( t_i \), we retrieve the $\text{top-}n$ most relevant papers to form the initial candidate set \( D = \{ a_1, a_2, \dots, a_n \} \). Then,  based on  \( D \), LLMs are prompted to first create a survey outline to avoid brief outputs from one-shot generation, and then expand each section in parallel to construct the final survey.

\textbf{Task 3: Human-guided Survey Generation:}  \label{task3} 
In this task, we remove the retrieval stage of RAG; instead, the survey is generated based on a gold-standard survey outline and selected references, both extracted from human-written surveys.  This setup simulates a realistic scenario in which authors, having already selected relevant literature and a predefined outline, can then focus on guiding LLMs to write the survey.

To provide publicly accessible input, the abstracts of the cited papers are used as the primary information in our study. The input and output for the three tasks are detailed in the Appendix \ref{sec:input_output}. 

\setlength{\tabcolsep}{5pt}
\begin{table*}[ht]
\centering
\small 
\resizebox{16cm}{!}  
{
\begin{tabular}{lrrrrrrrrr}
\Xhline{0.8pt}
\rowcolor{gray!15} 
\textbf{Dataset} & \textbf{Domains} & \textbf{\#Docs} & \makecell[r]{\#Input\\Len} & \makecell[r]{\#Target\\Len}& \makecell[r]{\#Input\\Docs} & \makecell[r]{Structural\\ Outline}& \makecell[r]{Quality\\Indicators} & \makecell[r]{Multi-level\\Citation} & \makecell[r]{For Survey\\Generation} \\ \hline
\midrule
\texttt{\textbf{PubMed}} (\citeyear{cohan2018discourse})         & Bio    & 133K  & 3016   & 203   & 1      & \cmark  & \xmark & \xmark & \xmark \\
\texttt{\textbf{ArXiv}} (\citeyear{cohan2018discourse})           & Mixed  & 215K  & 4938   & 220   & 1      & \cmark  & \xmark & \xmark & \xmark \\
\texttt{\textbf{SciSummNet}} (\citeyear{yasunaga2019scisummnet})       & CL     & 1K    & 4417   & 151   & 61.00  & \xmark   & \xmark & \xmark & \xmark \\
\texttt{\textbf{Multi-XScience}} (\citeyear{lu2020multi})  & CS     & 40.5K & 778    & 116   & 4.42   & \xmark   & \xmark & \xmark & \xmark \\
\texttt{\textbf{BigSurvey}}  (\citeyear{ijcai2022p591})      & Mixed  & 4.4K  & 11893  & 1051  & 76.30  & \xmark   & \xmark & \xmark & \xmark\\
\midrule
\texttt{\textbf{SciReviewGen}} (\citeyear{kasanishi2023scireviewgen})   & CS     & 10.2K & 12503  & 8082  & 68.00  &  \cmark & \xmark & \xmark & \cmark  \\
\rowcolor{blue!12}
\texttt{\textbf{SurveyGen(ours)}} & Mixed  & 4.2K  & 11423& 5115  & 57.58  & \cmark  & \cmark  & \cmark  & \cmark  \\
\bottomrule
\end{tabular}
}
\caption{Comparison with other scientific document summarization datasets. \texttt{\textbf{SurveyGen(ours)}} and SciReview \cite{kasanishi2023scireviewgen} are the only two suitable for survey generation. Compared to SciReview, our dataset further supplements all cited papers with quality indicators and second-level references, supporting more accurate document selection and citation network analysis. In addition,  \texttt{\textbf{SurveyGen}} includes surveys from multiple domains, such as Computer Science, Medicine, Biology, and Psychology, whereas SciReviewGen is limited to Computer Science.}
\label{tab:dataset-comparison}
\end{table*}

\subsection{\texttt{\textbf{SurveyGen}}: Dataset Construction} 
\label{sec:dataset-construction}

We developed \texttt{\textbf{SurveyGen}} based on S2ORC \cite{lo2020s2orc}, a large dataset containing 81.1 million English academic papers. In the preliminary search, we extracted articles by filtering titles that either contain  \textit{``a survey''}, \textit{``survey of''}, \textit{``a review''}, \textit{``literature review''}, \textit{``overview''} with full-text data available and publication years after 2010\footnote{Data collected from S2ORC up to December 2024.}. This resulted in a total of 8,676 candidate papers. 

Since title-based filtering may still include non-survey articles, we applied an additional filtering step using abstracts to further refine the candidate set. Specifically, inspired by previous work that LLMs are effective as NLI models for evaluating factual consistency \cite{gubelmann2023truth,chiang2024chatbot}, we prompted three LLMs to classify whether a candidate paper is a survey-type article based on its title and abstracts, following three criteria: \textbf{(1)} Explicit declaration of survey intent (e.g., \textit{``conducts a survey''} or \textit{``provides a survey''}). \textbf{(2)} Focus on survey papers, rather than proposing novel methodologies or experimental results. \textbf{(3)} Discussion of field-specific trends, challenges, or future directions. Papers without abstracts were excluded at this step. Based on these criteria, 6,851 out of 8,676 papers were identified as survey articles by a majority vote of the LLMs.

We then retrieved the full-text data of these surveys using their paper IDs from the S2ORC bulks\footnote{\url{https://api.semanticscholar.org/api-docs/}}. Here, the full-text data includes the \textbf{\underline{full body}} of the survey with \textbf{\underline{section divisions}}, as well as the \textbf{\underline{citation locations}} of each reference within the survey. This allows us to obtain the structural outline of each survey paper and map references to their corresponding sections, which serve as the key input for Task \hyperref[task3]{{3}}. At this point, we removed papers that had fewer than 30 references or fewer than three top-level sections, as they are too short to serve as meaningful surveys. Finally, we obtained 4,205 papers suitable for survey generation and constructed the \texttt{\textbf{SurveyGen}} dataset, which includes 115,376 sections,  242,143 references directly cited within the surveys, and 5,062,596 references cited by these cited papers.  

The data format is outlined in Appendix \ref{sec:dataformat}.  Table \ref{tab:dataset-comparison} compares  \texttt{\textbf{SurveyGen}} with other datasets for scientific document summarization.

\begin{figure*}[htbp]
    \centering
    \includegraphics[width=\textwidth, trim=5 0.5 0.4 2, clip]{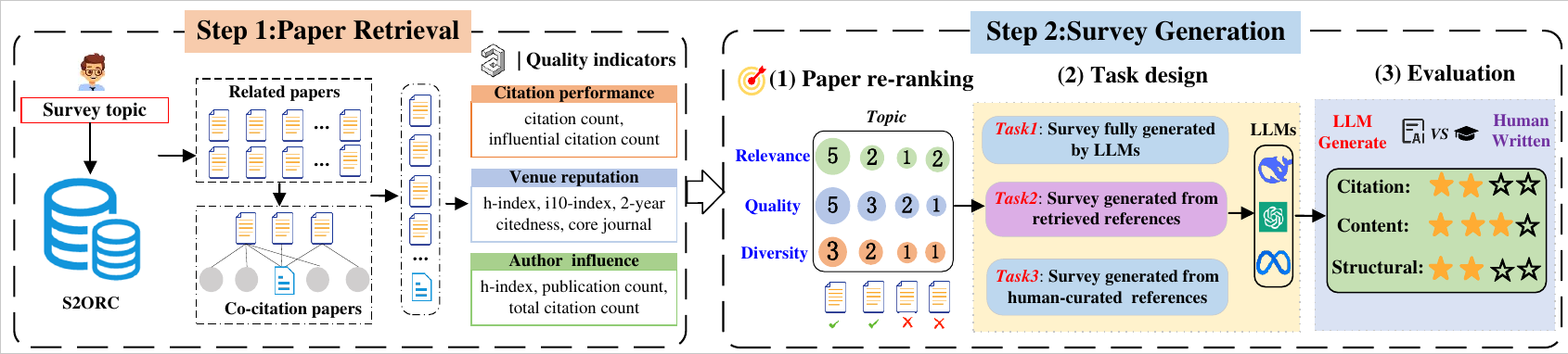}
    \caption{Overview of the QUAL-SG framework, which comprises two main stages: \textbf{paper retrieval} and \textbf{survey generation}. The retrieval stage includes three steps: \textbf{(1)} retrieving topic-relevant papers, \textbf{(2)}  expanding with frequently co-cited papers, and \textbf{(3)}  enriching them with quality-related metadata. Based on the retrieved set, the generation stage first re-ranks the papers from three key aspects, then prompts LLMs to perform tasks under different input conditions. Finally, we evaluate the generated surveys against human-written ones across multiple dimensions.}
    \label{fig:framework}
\end{figure*}

\subsection{Quality-Related Indicators Supplement}
\label{sec:Indicators-Supplement}
To facilitate citation-based evaluation, we first supplemented all survey papers and their directly cited references with basic metadata (e.g., abstract, DOI, publication venue, date, and research fields) from S2ORC, linked via corpus IDs. However, S2ORC does not provide sufficient metadata to measure the impact of academic papers. 

To address this limitation, we used DOIs of the involved papers to retrieve their corresponding metadata from the OpenAlex\footnote{\url{https://openalex.org/}} \cite{priem_2022_6936227} database and enriched them with additional quality-related signals. Specifically, we incorporated three well-known bibliometric indicators to measure the quality of scientific publications:  \textbf{(1) citation performance}: citation count and influential citation count\footnote{Citations identified by Semantic Scholar as impactful in context, rather than mere mentions in the bibliography.};  \textbf{(2) author influence}: h-index, publication count, and total citation count; and  \textbf{(3) venue reputation}: h-index, i10-index, and CORE status of the publication venue (journal or conference) \cite{hicks2015bibliometrics,donthu2021conduct}. As a result, each survey is now paired with its full text, complete with section divisions, and linked to its directly cited references. These references are further enriched with comprehensive multi-level quality indicators retrieved from OpenAlex, providing a robust foundation for evaluating generated surveys across multiple dimensions, such as citation accuracy, content quality, and structural consistency.

\textbf{Second-Level References Supplement:} In some cases, influential works relevant to a survey may not be semantically aligned with its primary topic. For example, when retrieving literature on \textit{``Deep Learning''}, seminal works such as the \textit{``Backpropagation algorithm''}  \cite{rumelhart1986learning} may obtain low semantic similarity scores as their titles and abstracts do not explicitly mention the topic. However, such papers are frequently cited by other retrieved references and are widely recognized as foundational to the field. To address this issue and support advanced citation network analysis, we further enriched the metadata for 5.06 million references cited by the papers referenced in all surveys. For each of these references, we extracted essential bibliographic details, including the title, abstract, DOI, and citation count.

\subsection{\texorpdfstring{QUAL-SG: \underline{\textsc{Qual}}ity-aware Literature Retrieval for \underline{\textsc{S}}urvey \underline{\textsc{G}}eneration}{QUAL-SG: Quality-aware Literature Retrieval for Survey Generation}}

\label{sec:QUAL-SG}

We propose QUAL-SG, a quality-aware extension of the naive RAG framework designed to improve the quality of retrieved literature for survey generation (Task \hyperref[task2]{{2}}). The overall framework is illustrated in Figure \ref{fig:framework}. As described in Section \ref{sec:task-design}, in the Naive-RAG framework, the survey topic is used as a query to retrieve relevant papers from external databases.  Formally, let \( q \) denote the topic derived from the human-written survey.  
Each candidate paper \( d_i \) in the external database is represented by its abstract embedding.  
We define the semantic similarity score as:
\vspace{-0.4em}
\begin{equation}
\text{Sim}(q, d_i) = \cos\left( \mathbf{v}_q, \mathbf{v}_{d_i} \right) = \frac{ \mathbf{v}_q \cdot \mathbf{v}_{d_i} }{ \| \mathbf{v}_q \| \, \| \mathbf{v}_{d_i} \| }
\end{equation}
where \( \mathbf{v}_q \) and \( \mathbf{v}_{d_i} \) are the embedding vectors of the query and the abstract of document \( d_i \), respectively.  

The $\text{top-}n$ with the highest embedding similarity scores are selected to form the initial candidate set $D = \{d_1, d_2, \dots, d_n\}$, where $n$ is set to exceed the number of references in the corresponding human-written survey to ensure sufficient candidate coverage.

Although the documents in \( D \) are topically relevant, certain papers may not exhibit strong semantic similarity to the query but still have a substantial impact within the research area (e.g., as seen in cases like \textit{``Backpropagation algorithm''} to \textit{``Deep Learning''}). Therefore, we expand \( D \) via a co-citation expansion: any paper cited by at least two papers in \( D \) is added to the set. Let $D_{ex}$ denote this expanded set.

Beyond topical relevance, crafting a high-quality survey also requires careful selection of cited papers \cite{paul2020art}. High-impact publications such as those published in reputable venues or frequently cited by other works generally contribute more significantly to the field \cite{kanellos2019impact}. Therefore, for each document in  \( D_{ex} \), we further collect a set of quality-related indicators, including citation performance, author influence, and venue reputation, as described in Section \ref{sec:dataset-construction}. 

 Then, we evaluate the quality of each candidate paper from three perspectives: \textbf{topical relevance}, \textbf{academic impact}, and \textbf{content diversity}. Specifically, for topical relevance, we employ LLMs-as-judge to assess the alignment between each candidate paper \( a_i \in D_{ex} \) and the survey topic \( t \). The relevance score is denoted as:
\begingroup
\setlength{\abovedisplayskip}{4pt}
\setlength{\belowdisplayskip}{2pt}
\begin{equation}
S_t = \text{LLM}_{\mathrm{judge}}(\bm{a}_i, t)
\end{equation}
\endgroup

For academic impact, we compute a weighted score that integrates three components: citation performance \( C(a_i) \), author influence \( A(a_i) \), and venue reputation \( V(a_i) \), since these factors are commonly associated with paper quality \cite{hicks2015bibliometrics, donthu2021conduct}. Each component is computed using a group-based scoring strategy, where raw indicator values are categorized into four ordinal levels based on percentile ranks. The overall academic impact score is defined as:
\begingroup
\setlength{\abovedisplayskip}{4pt}
\setlength{\belowdisplayskip}{3pt}
\begin{equation}
S_a = \alpha \cdot C(a_i)  + \beta \cdot A(a_i) + \gamma \cdot V(a_i)
\end{equation}
\endgroup
where \( \alpha \), \( \beta \), and \( \gamma \) are control variables that can be adjusted based on specific application needs.

For content diversity, we select papers that are topically relevant yet semantically distinct from others in the candidate pool to broaden the survey’s perspectives. To achieve this, we use the abstract of each paper as input and define the diversity of a candidate paper \( a_i \)  to a set of papers \( S \subseteq D_{ex} \) as the average semantic distance:
\begingroup
\setlength{\abovedisplayskip}{4pt}
\setlength{\belowdisplayskip}{2pt}
\begin{equation}
S_d(a_i, S) = \frac{1}{|S|} \sum_{a_j \in S} \text{Dist}(a_i, a_j)
\end{equation}
\endgroup

Finally, all candidate papers in \( D_{ex} \) are re-ranked based on their average ranks across  \( S_t \), \( S_a \), and \( S_d \). The $\text{top-}\mathcal{K}$ papers are selected to form the final set for survey generation, where $\mathcal{K}$ matches the number of references in the corresponding human-written survey to ensure a fair comparison.

\section{Experiments}
\subsection{Baselines}
\label{sec:baselines}
We selected three baselines for comparison.
\begin{itemize}[topsep=2pt, itemsep=2pt, parsep=0pt]
\item \textbf{Fully-LLMGen} \cite{tang2024llms}: Surveys are generated by LLMs based only on the given topic, without external inputs.
\item \textbf{Naive-RAG} \cite{wu2024automated}: Candidate papers are retrieved from an external literature database based on semantic similarity between the abstract and the survey topic. We use the same input fields as QUAL-SG to prompt LLMs for survey generation.
\item \textbf{Human-written}: The human-written surveys are from our \texttt{\textbf{SurveyGen}} dataset. 
\end{itemize}

For generation stages, we employed six LLMs as agents, including three  \textbf{Open-source LLMs:}  \textit{GLM-4-Flash} \cite{glm2024chatglm}, \textit{LLaMA-3.1-70B} \cite{meta2024llama31}, and \textit{DeepSeek-V3} \cite{liu2024deepseek}, and three \textbf{Closed-source LLMs:}  \textit{GPT-4.1-2025-04-14} \cite{openai2025gpt41}, \textit{Gemini-2.0-Flash} \cite{team2023gemini}, and \textit{Claude-3.7-Sonnet-20250219} \cite{anthropic2025claude37}. Implementation details are provided in Appendix~\ref{sec:implementation}.

To be cost-effective, our experiments are conducted on 120 highly cited surveys from \texttt{\textbf{SurveyGen}}, with 30 selected from each of four domains: Biology, Medicine, Psychology, and Computer Science. For Task \hyperref[task1]{{1}} and Task \hyperref[task3]{{3}}, we directly report the performance of different LLMs. For Task \hyperref[task2]{{2}}, we provide a comparative analysis between our QUAL-SG  and the baseline methods. A subset of survey examples is provided in Appendix ~\ref{sec:testdata}.

\subsection{Evaluation Metrics}
We consider human-written surveys as the ground truth for both automatic and human evaluations. 

\renewcommand{\arraystretch}{1}
\begin{table*}[ht]
\centering
\setlength{\tabcolsep}{5pt} 
\small 
\resizebox{15.5cm}{!}  
{

\begin{tabular}{l|cccc|ccc|cc}
\Xhline{1pt}
\rowcolor{gray!15}
\textbf{Model}
& \multicolumn{4}{c|}{
    \makecell{
      \textbf{Citation Quality} \\[2pt]
      \small \textbf{Acc. ↑} \quad \textbf{P ↑} \quad\hspace{3pt} \textbf{R ↑} \quad\hspace{1pt} \textbf{F1 ↑}
    }
}
& \multicolumn{3}{c|}{
    \makecell{
      \textbf{Content Quality} \\[2pt]
      \small \textbf{Sim. ↑} \quad \textbf{R-L ↑} \quad \textbf{KPR ↑}
    }
}
& \multicolumn{2}{c}{
    \makecell{
      \textbf{Structural Consistency} \\[2pt]
      \small \textbf{Rel$_\text{LLM}$} \quad \textbf{Overlap (\%)}
    }
}
\\
\hline

\midrule

\rowcolor{blue!12}
\multicolumn{10}{c}{{\small \faLockOpen \ } \textbf{Open-source LLMs}} \\
GLM-4-Flash         & 9.27 & 9.03 & 3.26 & 4.79
                  & 81.27 & \hspace{3pt}\underline{15.04} & 41.71 
                  & \hspace{3pt}2.44 & 10.62 \\
LLaMA-3.1-70B       & 15.43 & 11.48 & 2.74 & 4.42 
                  & \textbf{82.43} & \hspace{3pt}\textbf{15.36} & \underline{44.36}
                  & \hspace{3pt}\underline{2.62} & \underline{13.48} \\
DeepSeek-V3          & \underline{33.63} & 10.85 &  \underline{4.09} &  \underline{5.94}
                  & \underline{82.05} & \hspace{3pt}14.18 & 43.53 
                  & \hspace{3pt}2.57 & 11.03 \\
\midrule
\rowcolor{blue!12}
\multicolumn{10}{c}{{\small \faLock} \ \textbf{Closed-source LLMs}} \\
GPT-4.1            & 21.07 & \textbf{12.31} & 3.72 & 5.71
                  & 79.51 & \hspace{3pt}13.48 & 39.21 
                  & \hspace{3pt}2.39 & 10.95 \\
Gemini-2.0-Flash   & 22.20 & 8.97  & 3.59 & 5.13 
                  & 80.20 & \hspace{3pt}14.65 & 42.67 
                  & \hspace{3pt}2.50 & 12.39 \\
Claude-3.7-Sonnet  & \textbf{35.84} & \underline{11.79} & \textbf{5.78} & \textbf{7.76} 
                  & 81.32 & \hspace{3pt}13.77 & \textbf{46.59}
                  & \hspace{3pt}\textbf{2.65} & \textbf{14.89} \\

\bottomrule
\end{tabular}
}
\caption{Performance comparison of different LLMs on Task 1. ``Acc'' indicates whether the generated references are factually accurate and correspond to real papers. ``Sim'', ``R-L'', and ``KPR'' represent ``Semantic similarity'', ``ROUGE-L'', and ``Key Point Recall'', respectively. ``Rel$_\text{LLM}$'' represents structural consistency in LLM evaluations. The best results are marked \textbf{bold} and the second-best are \underline{underlined}.}
\label{tab:task1}
\end{table*}

\paragraph{Automatic evaluation:} 
The automatic evaluation includes three parts: \textbf{\textit{citation quality}}, \textbf{\textit{content quality}}, and \textbf{\textit{structural consistency}}. 
The formulas for the metrics and calculation details in this section are provided in the Appendix \ref{sec:metrics}.

\textbf{(1) Citation quality evaluation.} 
First, we assess how closely the references retrieved by RAG or generated by LLMs match those selected by humans. To address variations in title phrasing and formatting of the same article, we compute the textual similarity between each generated or retrieved reference and the human-selected ones. A reference is considered matched if the similarity exceeds a predefined threshold. We use precision, recall, and F1 score to evaluate citation overlap. Additionally, for Task \hyperref[task1]{{1}}, we compute citation accuracy to check whether the generated references are fabricated or hallucinated.

\textbf{(2) Content quality evaluation.} We first compute the semantic similarity between the LLM-generated and human-written surveys, and then report ROUGE\footnote{\url{https://pypi.org/project/pyrouge/}. All reported Rouge scores have a 95\% confidence interval in this paper.} score to quantify their textual overlap. Apart from semantic similarity evaluation,  we employ  Key Point Recall (KPR) \cite{qi2024long2rag,tang2024llms} to evaluate how effectively LLM-generated surveys capture the key points conveyed in human-written ones. 

\textbf{(3) Structural consistency evaluation.} In scientific writing, a well-structured survey typically features clear section divisions and coherent thematic development \cite{Wee03032016, paul2020art}. To evaluate structural consistency, we adopt two metrics: Overlap score and Relevance$_{\text{LLM}}$. Specifically, the Overlap score is defined as the number of sections between the generated and human-written surveys with semantic similarity exceeding a predefined threshold. Then, we prompt the LLM-as-judge to evaluate the structural alignment between LLM-generated and human-written surveys using a 5-point scale. 

\paragraph{Human evaluation:} 
Inspired by previous works \cite{kasanishi2023scireviewgen,liang2025surveyx}, we also conduct human evaluation to compare the LLM-generated and human-written surveys from the following four aspects: \textbf{\textit{topic relevance}}, \textbf{\textit{information coverage}}, \textbf{\textit{critical analysis}}, and \textbf{\textit{overall rating}}. The evaluation criteria and the detailed annotation process are provided in the Appendix \ref{sec:human evaluation}.

\section{Results and Analysis}

\subsection{Main Results}
\label{sec:main_results}
\renewcommand{\arraystretch}{1}
\begin{table*}[ht]
\centering
\setlength{\tabcolsep}{5pt} 
\small 
\resizebox{15.5cm}{!}  
{

\begin{tabular}{l|ccc|ccc|cc}
\Xhline{1pt}
\rowcolor{gray!15}
\textbf{Model}
& \multicolumn{3}{c|}{
    \makecell{
      \textbf{Citation Quality} \\[2pt]
      \small \hspace{4pt}\textbf{P ↑} \phantom{0.1} \hspace{5pt} \textbf{R ↑} \phantom{0.7} \hspace{4pt} \textbf{F1 ↑}    }
}
& \multicolumn{3}{c|}{
    \makecell{
      \textbf{Content Quality} \\[2pt]
      \small \textbf{Sim. ↑} \quad \textbf{R-L ↑} \quad \textbf{KPR ↑}
    }
}
& \multicolumn{2}{c}{
    \makecell{
      \textbf{Structural Consistency} \\[2pt]
      \small \textbf{Rel$_\text{LLM}$} \quad \textbf{Overlap (\%)}
    }
}
\\
\hline

\midrule

Fully-LLMGen      & 11.79 & 5.78 & 7.76 & 81.32 & 13.77 & 46.59 & \hspace{3pt}2.65 & 14.89 \\
Naive-RAG          & 5.18  & 6.94 & 5.93 & 82.37 & 12.90 & 42.17 & \hspace{3pt}2.43 & 12.22 \\
\textbf{QUAL-SG (Ours) }    & \textbf{15.87\textsuperscript{†}} & \textbf{17.71\textsuperscript{†}} & \textbf{16.73\textsuperscript{†}}
& \textbf{83.10\textsuperscript{†}} & \textbf{15.17\textsuperscript{†}} & \textbf{50.25\textsuperscript{†}}
& \hspace{4pt}\textbf{2.81\textsuperscript{†}} & \textbf{24.76\textsuperscript{†}} \\

\bottomrule
\end{tabular}
}

\caption{Performance of different models on Task 2. For Fully-LLMGen \cite{tang2024llms}, we directly report the results from Task 1. In the Naive-RAG setting \cite{wu2024automated}, retrieval is based on the semantic similarity between the survey topic and candidate abstracts.  Claude-3.7-Sonnet is used as the backbone for all methods. The best results are marked \textbf{bold}. † denotes significant differences to baselines ($p$-value $<$ 0.001).}
\label{tab:task2}
\end{table*}

\paragraph{Results for Task 1:} We report the evaluation results of different LLMs on Task \hyperref[task1]{{1}}. As shown in Table \ref{tab:task1}, Claude 3.7-Sonnet achieves the best overall performance across citation quality, KPR, and structural consistency. In content evaluation, LLaMA-3.1-70B achieves the highest similarity to human-written surveys (82.43\%) and the highest ROUGE-L (15.36\%). However, citation accuracy remains a major limitation: the best-performing model achieves only 35.84\%, indicating that \textbf{relying solely on LLMs for survey generation is insufficient for ensuring reliable reference generation}. Furthermore, compared to human-written surveys, although the LLM-generated content is semantically similar, it still shows significant gaps in key point coverage (46.59\%) and structural overlap (14.89\%). Lastly, closed-source and open-source LLMs exhibit distinct strengths: closed-source models consistently surpass open-source models in citation quality and structural consistency, while open-source models deliver comparable results in content generation.

\paragraph{Results for Task 2:} Table \ref{tab:task2} summarizes the results of  different models on Task  \hyperref[task2]{{2}}. Compared with the Fully-LLMGen approach, the Naive-RAG method, despite retrieving authentic literature from external databases, yields the lowest citation quality.  In contrast,  \textbf{our proposed QUAL-SG  achieves the highest citation quality} (F1 score of 16.73\%), outperforming Naive-RAG and Fully-LLMGen by 10.80\% and 8.97\%, respectively. QUAL-SG also surpasses both baselines in content quality (Similarity +0.73\%, ROUGE-L +1.40\%, KPR +3.66\%) and structural consistency (LLM evaluation +0.16 on a 5-point scale, semantic overlap +12.54\%).

The results suggest that while the Naive-RAG framework can improve the factual accuracy of generated references, it remains limited in identifying truly human-preferred or high-quality references from the large-scale academic database. In contrast, QUAL-SG mitigates this limitation via a re-ranking module that integrates topical relevance, academic impact, and content diversity, yielding reference selections better aligned with human preferences. This improvement in citation quality, in turn, enhances the overall content quality and structural consistency of the generated surveys.
\begin{figure*}[!htb]
  \centering
  \begin{subfigure}[t]{0.24\textwidth}
    \centering
    \includegraphics[width=\linewidth,trim={0 8 2 4},clip]{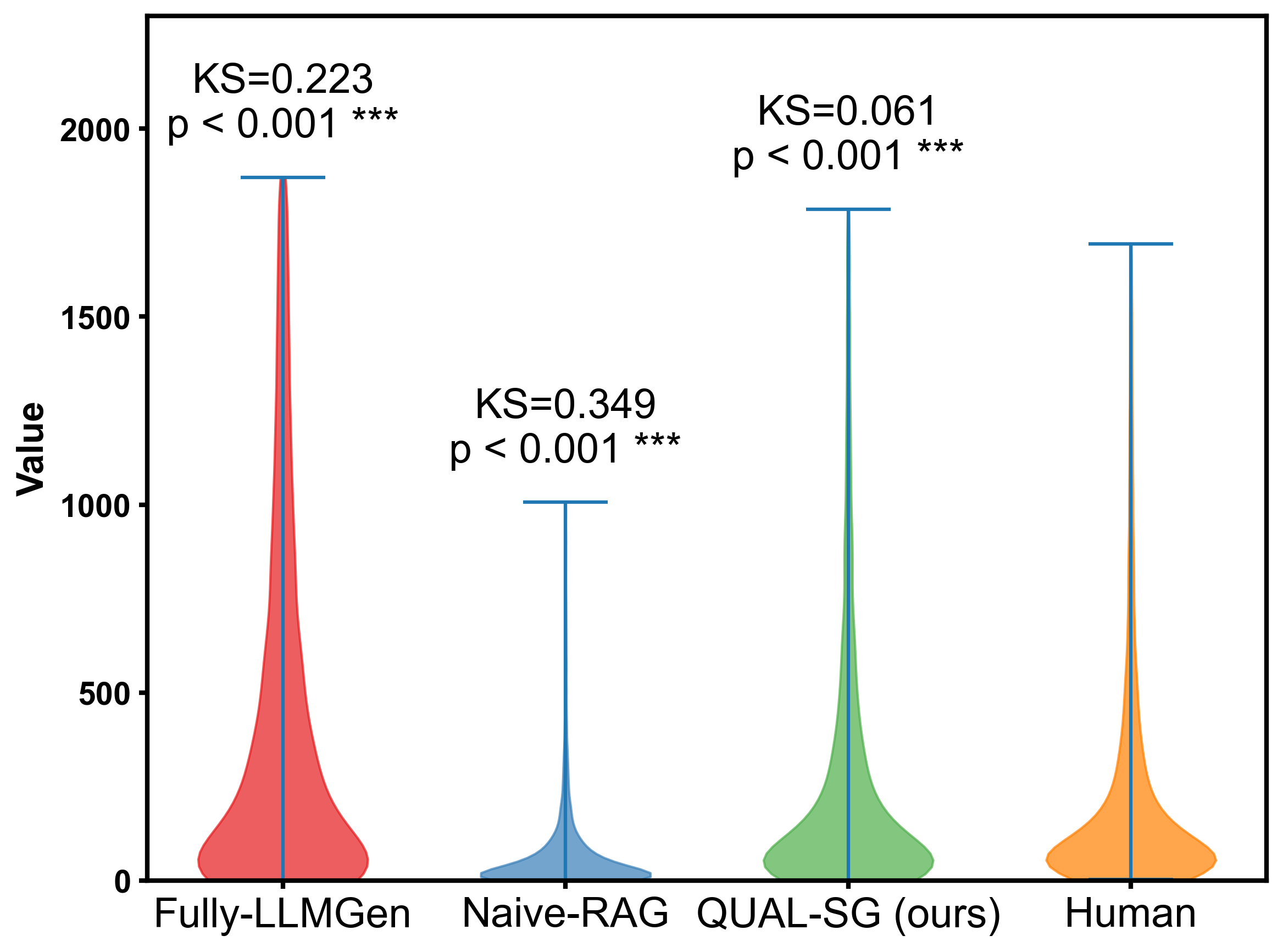}
    \caption{Citation Count}
    \label{fig:3a}
  \end{subfigure}\hfill
  \begin{subfigure}[t]{0.24\textwidth}
    \centering
    \includegraphics[width=\linewidth,trim={0 8 0 4},clip]{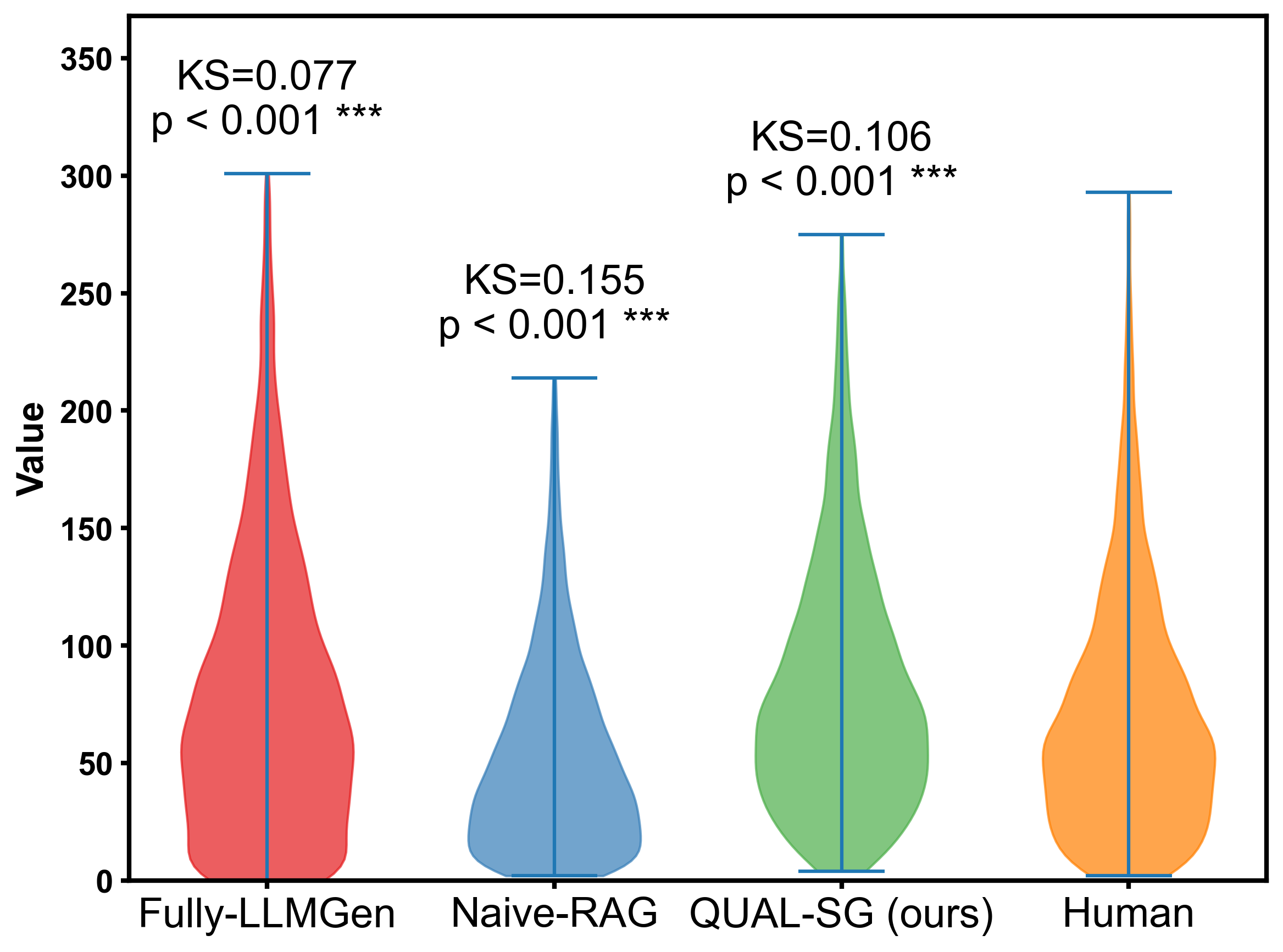}
    \caption{Author H-Index}
    \label{fig:3b}
  \end{subfigure}\hfill
  \begin{subfigure}[t]{0.24\textwidth}
    \centering
    \includegraphics[width=\linewidth,trim={0 8 0 4},clip]{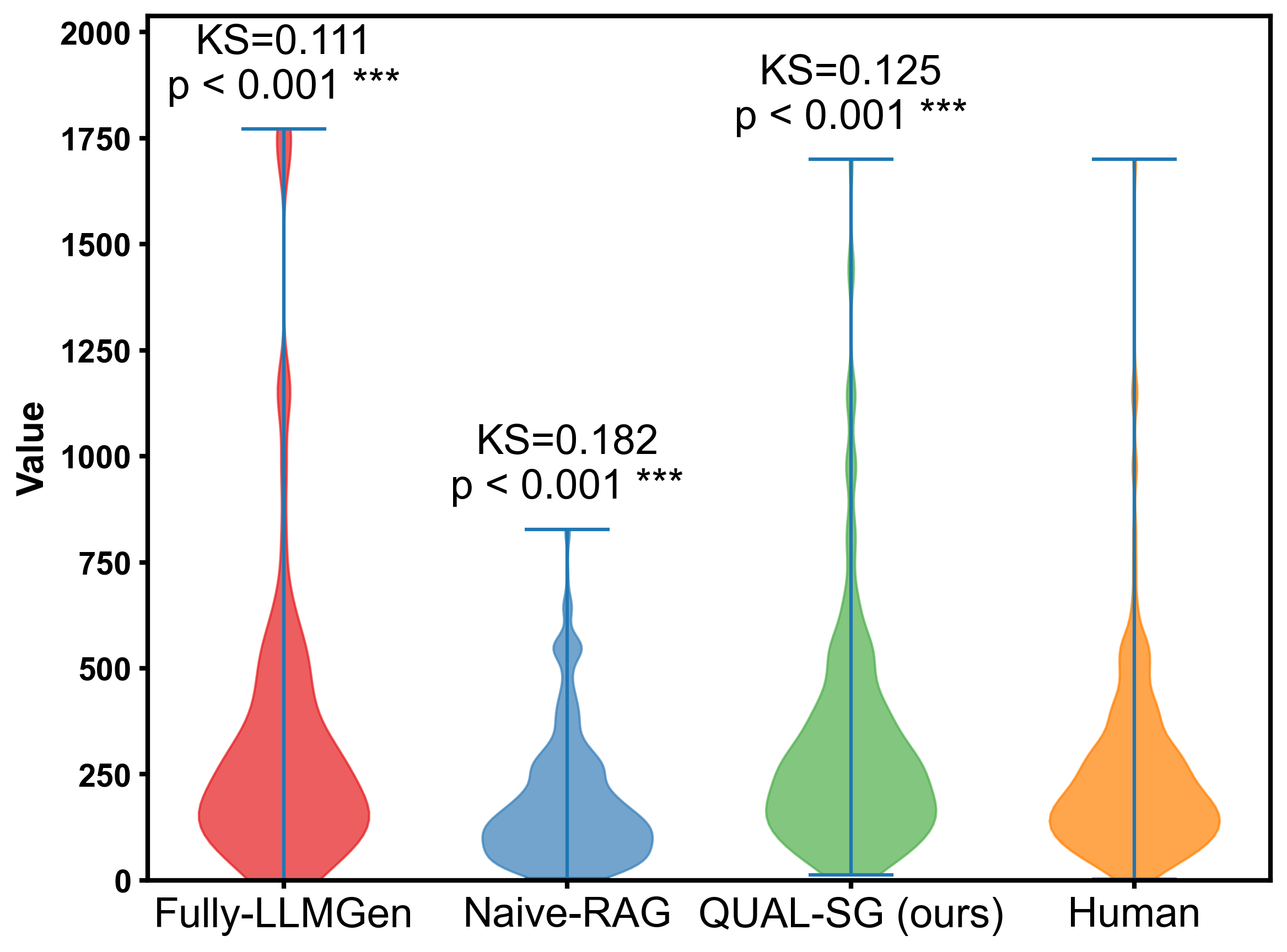}
    \caption{Venue H-Index}
    \label{fig:3c}
  \end{subfigure}\hfill
  \begin{subfigure}[t]{0.24\textwidth}
    \centering
    \includegraphics[width=\linewidth,trim={0 8 0 4},clip]{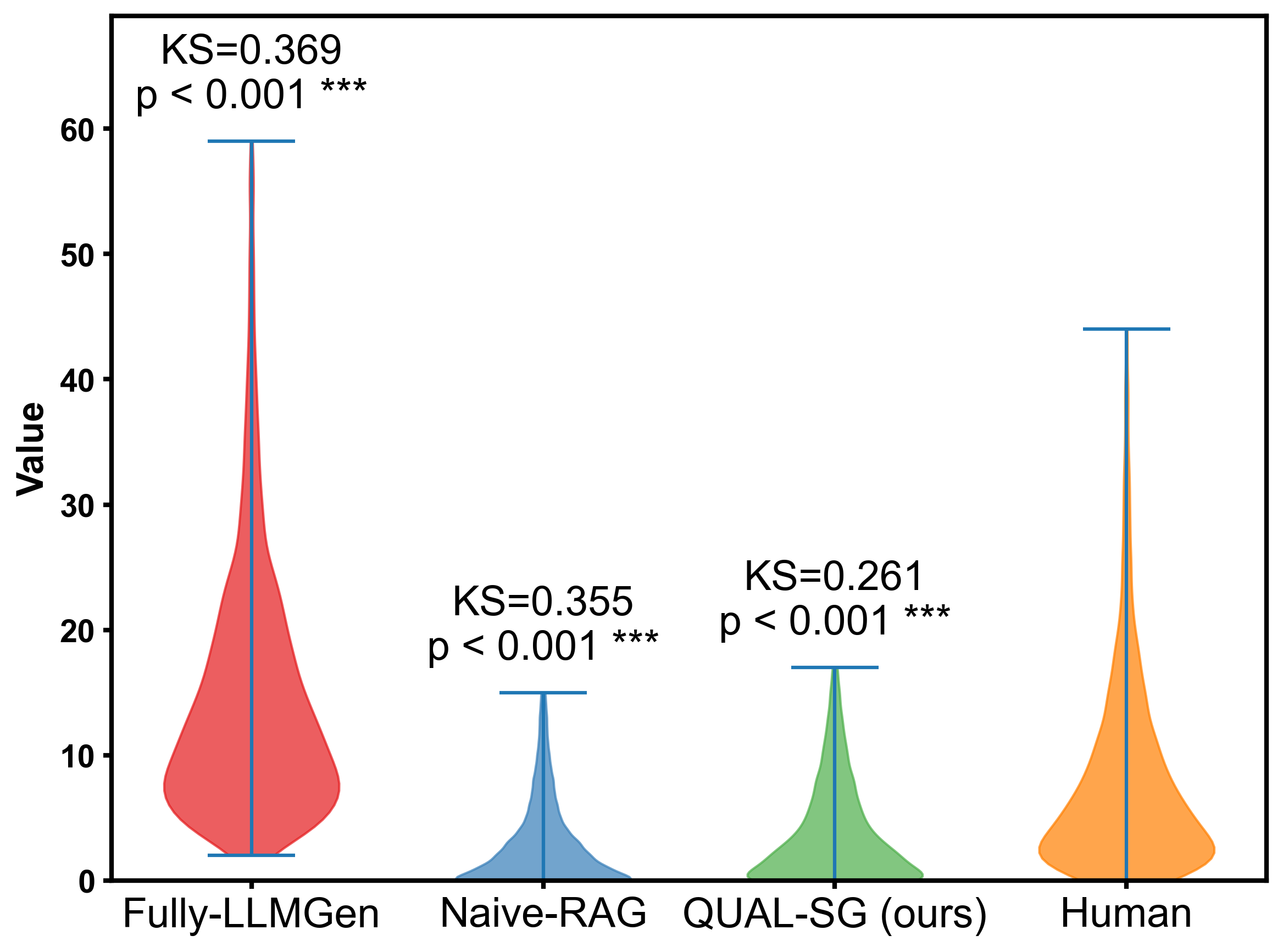}
    \caption{Years Gap}
    \label{fig:3d}
  \end{subfigure}
  \caption{Comparison of reference selection distributions across models. “KS” denotes the Kolmogorov–Smirnov statistic against the human baseline (lower values indicate closer alignment), “\textit{p}” is the associated p-value, and “Years Gap” denotes the difference in publication years between the reference and the survey. For Fully-LLMGen, the survey year is set to 2025. Claude-3.7-Sonnet is used as the backbone LLM for all methods.}
  \label{fig:3}
\end{figure*}

\paragraph{Results for Task 3:} For Task \hyperref[task3]{{3}}, since both the candidate references and the outline are directly extracted from human-written surveys, we only report the content evaluation results of different LLMs, as shown in Table \ref{tab:task3}. We observe that  \textbf{when LLMs are provided with more accurate references and outlines, their generated content quality improves accordingly} compared to Task \hyperref[task2]{{2}}, which involves no human intervention. Among the models, the open-source LLaMA-3.1-70B still achieves the highest content similarity (84.39\%) and ROUGE-L (17.16\%), while Claude-3.7-Sonnet obtains the highest KPR (54.67\%). Overall, with human intervention, open-source models exhibit a strong capability to compete with advanced closed-source models in the survey generation task.

\renewcommand{\arraystretch}{1.1}
\begin{table}[t] 
\centering
\setlength{\tabcolsep}{6pt}
\small 
\resizebox{7cm}{!}  
{
\begin{tabular}{lccc}
\Xhline{1pt}
\rowcolor{gray!15}
\textbf{Model} & \hspace{3pt}\textbf{Sim. ↑} & \hspace{3pt}\textbf{R-L ↑} & \hspace{3pt}\textbf{KPR ↑} \\
\hline
\hline

\rowcolor{blue!12}
\multicolumn{4}{c}{{\small \faLockOpen \ } \textbf{Open-source LLMs}} \\
GLM-4-Flash        & 82.04 & \underline{16.29} & 46.88 \\
LLaMA-3.1-70B        & \textbf{84.39} & \textbf{17.16} & \underline{52.13} \\
DeepSeek-V3           & \underline{83.97} & 15.25 & 49.50 \\
\midrule
\rowcolor{blue!12}
\multicolumn{4}{c}{{\small \faLock} \textbf{Closed-source LLMs}} \\
GPT-4.1            & 82.59 & 13.82 & 50.02 \\
Gemini-2.0-Flash   & 83.74 & 15.62 & 51.76 \\
Claude-3.7-Sonnet  & 84.22 & 15.43 & \textbf{54.67} \\
\bottomrule
\end{tabular}
}
\caption{Content quality evaluation results of different LLMs on Task 3. The best results are marked \textbf{bold} and the second-best are \underline{underlined}.}
\label{tab:task3}
\end{table}

\subsection{Further Analysis of Reference Selection}
We further analyze the distribution of references yielded by different models in Task \hyperref[task2]{{2}}, as shown in Figure \ref{fig:3}. The results indicate that QUAL-SG exhibits the closest alignment to human-written surveys in citation count and temporal distribution of selected references, and achieves competitive performance in author H-index and venue H-index (Figure~\ref{fig:3a}~\textasciitilde~\ref{fig:3d}). Specifically, Fully-LLMGen exhibits a pronounced long-tail distribution in reference selection, with most selected papers concentrated in the less cited studies. The poor performance of Naive-RAG highlights the limitation of purely semantic retrieval, as many retrieved papers,  although semantically relevant, fail to meet the quality standards expected for survey writing. Regarding the temporal distribution, human-written surveys tend to favor papers published within the preceding decade, while Fully-LLMGen often overlooks recent studies due to outdated training data.

\begin{figure*}[!htb]
  \centering
  \setlength{\tabcolsep}{0.2pt}        
  \renewcommand{\arraystretch}{0.9}  

  \begin{tabular}{cccc}
    \subcaptionbox{Similarity\label{fig:sub1}}{%
      \includegraphics[width=\dimexpr 0.25\linewidth-2pt\relax]{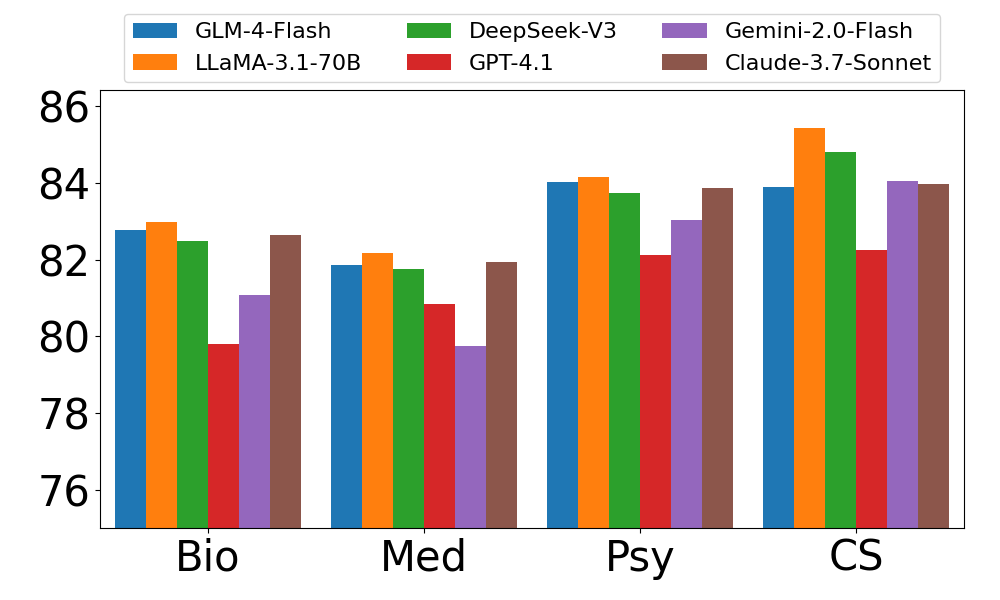}}%
    &\subcaptionbox{Rouge-L\label{fig:sub2}}{%
      \includegraphics[width=\dimexpr 0.25\linewidth-2pt\relax]{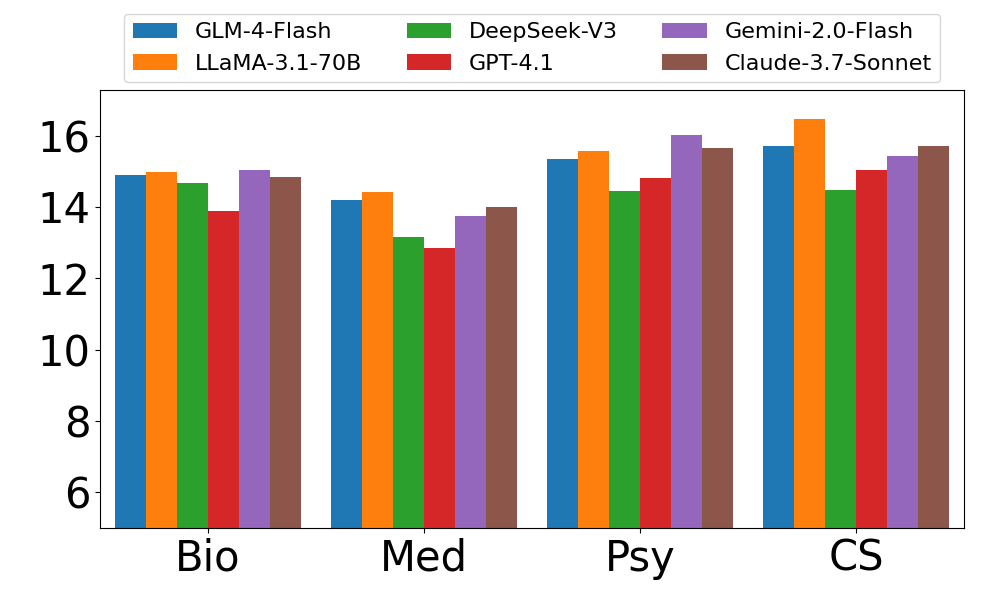}}%
    &\subcaptionbox{KPR\label{fig:sub3}}{%
      \includegraphics[width=\dimexpr 0.25\linewidth-2pt\relax]{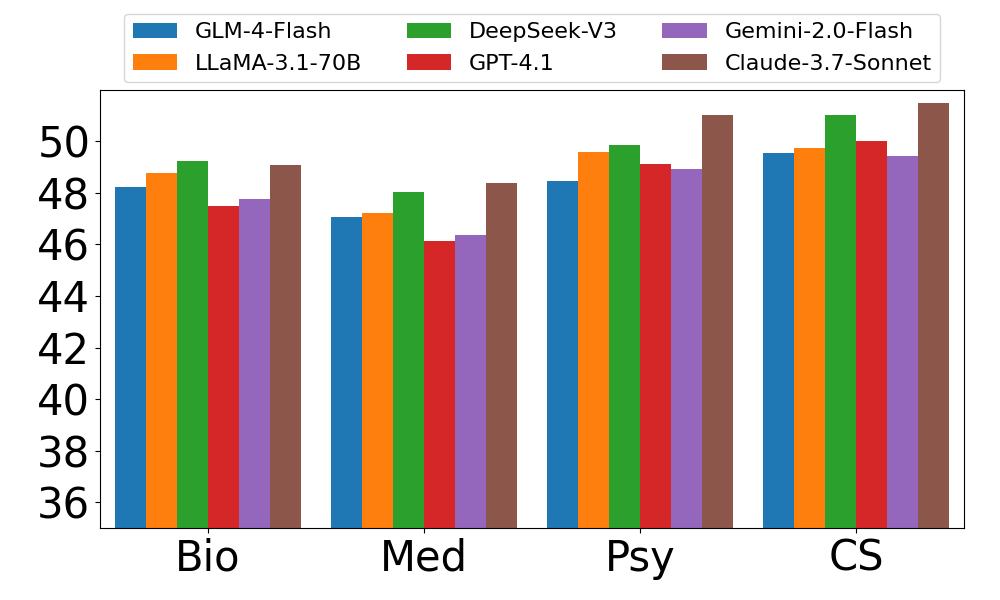}}%
    &\subcaptionbox{Structural Overlap\label{fig:sub4}}{%
      \includegraphics[width=\dimexpr 0.25\linewidth-2pt\relax]{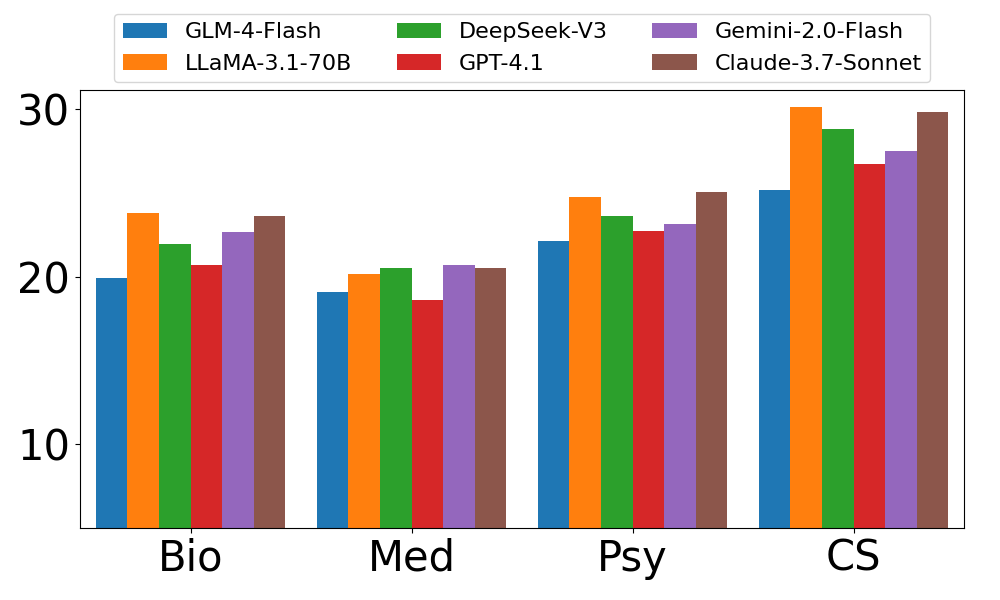}}%
  \end{tabular}

  \caption{Performance comparison of different models across disciplines on Task 2. “Bio”, “Med”, “Psy”, and “CS” denote Biology, Medicine, Psychology, and Computer Science, respectively. “KPR” refers to Key Point Recall.}
  \label{fig:figure4}
\end{figure*}

\subsection{Cross-Disciplinary Comparison of LLMs} 
We extend the analysis from Task \hyperref[task2]{{2}} to compare the performance of each LLM powering QUAL-SG across academic disciplines. As shown in Figure \ref{fig:sub1},  the models yield relatively stable content similarity across domains. This observation is further confirmed by one-way ANOVA tests conducted for each model, which reveal no statistically significant differences across disciplines: GLM-4-Flash (\textit{p}=0.17), DeepSeek-V3 (\textit{p}=0.42), LLaMA-3.1-70B (\textit{p}=0.31), GPT-4.1 (\textit{p}=0.21), Gemini-2.0-Flash (\textit{p}=0.39), and Claude-3.7-Sonnet (\textit{p}=0.32).

We then report the ROUGE-L scores for different LLMs across disciplines. As shown in Figure~\ref{fig:sub2}, scores in Computer Science and Psychology are generally higher than those in Medicine and Biology, with LLaMA-3.1-70B consistently outperforming other models. Moreover, all models exhibit statistically significant performance differences across disciplines (\textit{p}<.001).

Similarly, KPR scores (Figure~\ref{fig:sub3}) follow the same trend, with higher scores in Computer Science and Psychology across all models. Claude-3.7-Sonnet consistently achieves the best KPR score. However, the differences across disciplines are not statistically significant for individual models: GLM-4-Flash (\textit{p}=0.12), DeepSeek-V3 (\textit{p}=0.36), LLaMA-3.1-70B (\textit{p}=0.25), GPT-4.1 (\textit{p}=0.40), Gemini-2.0-Flash (\textit{p}=0.27), and Claude-3.7-Sonnet (\textit{p}=0.33).

For structural consistency (Figure~\ref{fig:sub4}), LLaMA-3.1-70B achieves the best performance in Computer Science, Biology, and Psychology, while Gemini-2.0-Flash leads in Medicine. All models show statistically significant differences in structural consistency across disciplines (\textit{p}<.001).

\subsection{Comparison with Other Ranking Models}
We compare our QUAL-SG with UPR \citep{sachan-etal-2022-improving} and RankGPT \citep{sun-etal-2023-chatgpt}, both designed for ranking candidates in the RAG pipeline.  Since the generation stage mainly depends on the selected references as sources, we report their performance only in the retrieval stage, as this more directly reflects the impact of candidate ranking.  As shown in Table \ref{tab:ranking_results}, our method outperforms UPR, which relies on probability-based token-level ranking. While RankGPT incorporates this criterion through its instructions, QUAL-SG employs a more direct strategy through weighted aggregation, demonstrating greater robustness when handling multiple ranking criteria. 

\begin{table}[h]
\centering
\resizebox{\linewidth}{!}{ 

\begin{tabular}{lccc}
\Xhline{1pt}
\rowcolor{gray!15}
\textbf{Model} & \textbf{P\%↑} & \textbf{R\%↑} & \textbf{F1\%↑} \\
\hline
\hline

UPR \citep{sachan-etal-2022-improving} & 10.28 & 10.63 & 10.45 \\
RankGPT \citep{sun-etal-2023-chatgpt} & \underline{14.55} & \underline{15.09} & \underline{14.81} \\
\textbf{QUAL-SG (ours)} & \textbf{15.87} & \textbf{17.71} & \textbf{16.73} \\
\bottomrule
\end{tabular}
}
\caption{Citation quality comparison of different ranking models. For RankGPT, we prompt it to rank papers according to the same three criteria (\sref{sec:QUAL-SG}) used in our QUAL-SG. The best results are marked \textbf{bold} and the second-best are \underline{underlined}.}
\label{tab:ranking_results}
\end{table}

\subsection{Human Evaluation Results}
\label{sec:human_results}
The human evaluation results are presented in Table \ref{tab:human} in the Appendix \ref{sec:human evaluation}. We can observe that Task  \hyperref[task3]{{3}} is generally rated as more acceptable by human evaluators. This highlights the importance of key preprocessing steps, such as high-quality reference selection and effective outline construction, in guiding LLMs to generate more reliable scientific surveys. However, despite the comparable performance in terms of topic relevance,  the generated surveys currently \textbf{\textit{fail to provide sufficient information coverage and critical analysis}}.

\section{Discussion and Future Directions}
\label{sec:discussions-future}
\paragraph{LLM for Automatic Survey Generation: Are We There Yet?} The results in Section \ref{sec:main_results} indicate that neither Fully LLM-based nor RAG-based approaches have achieved human-level performance. As highlighted in \cite{liang2025surveyx,tang2024llms}, hallucinated information, such as fabricated references and factual inaccuracies, remains a critical challenge in LLM-generated surveys. Although RAG-based methods reduce hallucinations by retrieving external sources, the retrieved papers are often only topically relevant and misaligned with human preferences. While LLMs have demonstrated efficiency and the ability to generate content considered useful by human evaluators \cite{wang2024autosurvey}, our human evaluation results (\sref{sec:human_results}) reveal that, despite strong topical relevance, LLM-generated surveys exhibit limited coverage and in-depth analysis, both essential for high-quality scientific surveys. Therefore, while LLMs can assist in survey generation, they are still unable to independently craft surveys that meet academic standards at the current stage.

\paragraph{Future Directions for Enhancing Survey Generation} As shown in Section \ref{sec:main_results}, quality-based ranking of candidate references effectively improves the citation performance of generated surveys. This can be further enhanced through several strategies. For example, citation network analysis can be employed to capture the global relationships among papers and identify influential studies. Additionally,  analyzing human citation behavior—such as citation intent, frequency, and location in the textual context—can inform better reference selection mechanisms.  Training reference selection models on human-annotated datasets is also a potential option for collecting literature suitable for survey generation. On the generation stage, relying solely on abstracts as input significantly limits the information coverage, as it fails to fully capture the paper's broader details.  Future work could leverage full-text information to enable more comprehensive contextual understanding, as well as explore human-in-the-loop discourse control, factual consistency verification, and advanced long-document modeling to improve survey quality.

\paragraph{Real-World Applicability and Deployment}
Our framework is designed with modular components, including embedding-based retrieval, co-citation expansion, and re-ranking, which can be parallelized or extended. For the retrieval stage, we choose S2ORC \citep{lo2020s2orc} as the external database because its papers are peer-reviewed and have full dataset downloads, which can be stored locally and used for a one-time embedding computation. In practice, it can be replaced with other sources such as arXiv\footnote{\url{https://arxiv.org/}} or PubMed\footnote{\url{https://pubmed.ncbi.nlm.nih.gov/}}, depending on user needs.  Additionally, numerous well-established embedding models are available on the MTEB leaderboard\footnote{\url{https://huggingface.co/spaces/mteb/leaderboard}}, offering a range of trade-offs between accuracy, model size, and computational efficiency. For the co-citation expansion module, we rely on the OpenAlex \citep{priem2022openalexfullyopenindexscholarly} database for citation analysis. OpenAlex also provides free APIs and allows bulk download of citation data. Similarly, users can replace OpenAlex with other citation databases, such as Scopus \cite{scopus_api} and SciSciNet \citep{lin2023sciscinet}. As for the re-ranking, we assume it is highly adaptable to different downstream needs. Since we will release the \texttt{\textbf{SurveyGen}}, users can customize re-ranking strategies according to their specific preferences. In the generation stage, the results (\sref{sec:main_results}) show that open-source models (e.g., LLaMA-3.1-70B) can achieve competitive performance compared to closed-source commercial LLMs such as GPT-4.1. This offers users greater flexibility based on their budget, deployment needs, and infrastructure.

\section{Conclusion}
We introduce \texttt{\textbf{SurveyGen}}, a new dataset designed to support scientific survey generation. Building on this resource, we propose QUAL-SG, an enhanced RAG framework that improves upon Naive-RAG by identifying higher-quality references during literature retrieval. Experimental results show that QUAL-SG outperforms semantic similarity-based RAG methods across key aspects, including citation quality, content quality, and structural consistency of the generated surveys. Finally, we conduct a human evaluation to assess the impact of human intervention at different stages of the survey generation process. Our findings show that providing more accurate references and a well-structured outline enables LLM to generate surveys more aligned with human-written ones; however, there remains considerable room for improvement in both citation and content quality to meet human expectations.

\section*{Limitations}
\label{sec:limitations}

We acknowledge several limitations in our work. 

\paragraph{Input Limitation.} For copyright reasons, our approach is restricted to using only abstracts and bibliographic metadata of the retrieved papers, without access to full-text content. This limitation may hinder the LLM’s ability to capture finer-grained details and structural elements that are often present in full-length papers. Hence, the generated surveys may lack depth and completeness compared to human-written surveys that draw on the entire papers.

\paragraph{Post-generation Refinement.} To reduce API call costs, we did not perform post-generation refinement to the LLM outputs, such as language polishing, citation formatting, or structural adjustments. These post-processing steps could further improve the personalization and overall quality of the generated surveys. Also, our work focuses on generating textual survey content and does not include visual elements such as figures, tables, or diagrams, which are often present in published scientific surveys. Lastly, for longer surveys, models like \textit{Claude-3-Haiku}\footnote{\url{https://www.anthropic.com/news/claude-3-haiku}} may offer superior performance due to their extended context handling capabilities. 

\paragraph{Data Contamination.} We acknowledge the possibility of data contamination, as some surveys or key references (\sref{sec:baselines}) used are open access and may have been included in the training data of the LLMs, potentially leading to slightly different performance estimates. Although we do not explicitly control for this factor in our evaluation process, such contamination is a general challenge in benchmarking LLMs on open-domain generation tasks \cite{xu2024benchmarkingbenchmarkleakagelarge}. Moreover, since all baselines in our study are based on mainstream LLMs, any potential contamination would be shared and thus unlikely to impact the relative comparison.

\paragraph{Evaluation Sample Scope.} While our empirical evaluation focuses on a subset of 120 relatively short surveys spanning multiple disciplines—selected to balance cost and feasibility in \emph{academic settings}—we expect similar performance trends to hold across the full dataset. We encourage the broader research community to further benchmark models using our dataset and framework to extend our findings across broader contexts.

\section*{Ethics Statement}
\label{sec:ethics}

\paragraph{Data Collection, Ethics, and Licensing.} 
Our \texttt{\textbf{SurveyGen}} dataset is constructed based on S2ORC \cite{lo2020s2orc}, a large corpus of scientific papers released under the CC BY-NC 4.0\footnote{https://creativecommons.org/licenses/by-nc/4.0/deed.en}. The dataset includes metadata extracted from the papers, such as author names, venue names, citation counts, and h-index values. No sensitive personal data (e.g., contact details or affiliations) is included. All metadata was collected in compliance with the terms of their sources and is used strictly for non-commercial academic research. The dataset is not intended for ranking or evaluating individuals or venues. We are committed to handling the data responsibly and ethically and will release our dataset under the same non-commercial license to ensure transparency and responsible data usage.

\paragraph{Caution about Use of LLMs.} While our QUAL-SG framework leverages LLMs to generate scientific surveys and strives to maintain the factual accuracy of the literature, there remains a concern of factual inconsistencies during the generation process. We advise users to critically evaluate the generated content, especially when using it for subsequent scientific research or practical applications. The LLM-generated survey is for reference only and should not be regarded as a substitute for peer-reviewed articles or expert judgment.
\section*{Acknowledgements}
\label{sec:acknowledgements}
 This work is supported by the National Natural Science Foundation of China (No.72074113) and the Natural Sciences and Engineering Research Council of Canada (NSERC). We gratefully acknowledge the Digital Research Alliance of Canada (CCDB) for providing GPU resources. Mir Tafseer Nayeem is supported by a Huawei PhD Fellowship. We thank Yi Zhao, Heng Zhang, Wenqing Wu, and the anonymous reviewers for their valuable feedback. Tong Bao also thanks his parents and his girlfriend (S. Song) for supporting him during his visit to the University of Alberta.
\bibliography{custom}
\clearpage
\appendix
\twocolumn[{%
 \centering
 \Large\bf Supplementary Material: Appendices \\ [20pt]
}]

\section{Related Work}
\label{sec:rel-work}

\paragraph{Dataset for Scientific Literature Summarization:}  While scientific literature summarization has been extensively studied, most available datasets are limited to single-document scenarios. For instance, SciTLDR \cite{cachola2020tldr} contains both author-written and expert-derived summaries for scientific paper summarization tasks. \citet{cohan2018discourse} introduced a dataset from PubMed and arXiv for long document summarization. However, real-world scientific writing often integrates insights from multiple studies, which requires multi-document summarization datasets. To address this,  \citet{lu2020multi} proposed Multi-XScience, which extends single-document summarization by incorporating multiple source papers to generate a cohesive summary. \citet{deyoung2021msˆ2} proposed MS² for summarizing multiple medical studies to generate comprehensive surveys.

The work most similar to ours is SciReviewGen \cite{kasanishi2023scireviewgen}, which created a dataset of over 10,000  surveys in the computer science domain with cited references within the surveys. Our dataset differs in that \texttt{\textbf{SurveyGen}} additionally provides extensive metadata for all the referenced papers for evaluation purposes, including bibliographic information for papers (e.g., title, abstract, topics), citation performance (e.g., citation count, influential citation count), author-level influence indicators (e.g., publication count, h-index, and total citations), and venue-level reputation metrics (e.g., h-index, mean-citedness, i10-index). Unlike SciReviewGen, which focuses primarily on survey generation, our \texttt{\textbf{SurveyGen}} offers a more comprehensive benchmark for assessing citation reliability, content quality, and structural alignment in LLM-generated surveys. Finally, \texttt{\textbf{SurveyGen}} also supports the evaluation of surveys across multiple disciplines, while SciReviewGen is limited to computer science.

\paragraph{Automatic Literature Survey Generation with LLMs:}  While LLMs have demonstrated impressive performance in text generation tasks, generating content that meets the accuracy, structure, and logical coherence required for scientific surveys remains a challenge \cite{tang2023evaluating, lehr2024chatgpt, ELBADAWI2024123741}.To address this issue, some studies integrate RAG techniques with LLMs and define output templates to control the structure of the generated content \cite{lai2024instruct,agarwal2024llms,tang2024llms}. For instance, \citet{wang2024autosurvey} proposed AutoSurvey, which employs a two-stage generation strategy: first, retrieving relevant literature to generate a detailed outline, and then drafting individual sections and integrating them into a cohesive review. Similarly, \citet{liang2025surveyx} introduced SurveyX, which employs online reference retrieval to gather relevant literature and utilizes a pre-processing method called AttributeTree to extract and organize key information from these sources. \citet{wu2024automated}  implemented a multi-layered quality control strategy to mitigate hallucination issues during the literature review generation process. While the above studies provide valuable insights into this task, our work offers more reliable sources, improved retrieval strategies, and a more rigorous evaluation against human-written surveys to explore the upper limits of LLMs.

\section{Input and Output Settings}
\label{sec:input_output}

The input and output texts for the three tasks are as follows:

\textbf{Task \hyperref[task1]{1}}: The LLMs are provided only with the survey topic. They are first prompted to generate a structured outline along with brief descriptions for each section, and then to produce the full survey content based on that outline. 

\textbf{Task \hyperref[task2]{2}}: During the retrieval stage, the survey topic is used as a query to retrieve relevant literature from external databases. In the generation stage, the input includes the survey topic, along with the titles, abstracts, and quality-related metadata of the retrieved papers. The generation process follows the same steps as in Task \hyperref[task1]{1}, where the LLMs are instructed first to generate an outline with section descriptions and then write the corresponding content for each section to form the final survey.

\textbf{Task \hyperref[task3]{3}}: In this task, all references are sourced from human-written surveys, and the bibliographic information provided for each reference is consistent with that used in Task 2. In addition, we provide the outline of each human-written survey, with all cited references grouped under their corresponding sections. The LLMs are then instructed to generate each section using the selected references and the corresponding outline information.

\section{Implementation Details}
\label{sec:implementation}

During the literature retrieval stage, we utilize \textbf{Semantic Scholar}\footnote{\url{https://api.semanticscholar.org/api-docs/datasets}} as the external literature database, and use the \textbf{bge-large-en-v1.5}\footnote{\url{https://huggingface.co/BAAI/bge-large-en-v1.5}} as the embedding model to compute semantic similarity throughout the pipeline. To mitigate potential self-evaluation bias,  \textbf{GPT-4o} \cite{openai2024gpt4o} is selected as the LLM agent for evaluation purposes.

To implement our QUAL-SG framework, we first use the S2ORC API\footnote{\url{https://api.semanticscholar.org/graph/v1/paper/search}} to retrieve the 300  papers most relevant to the survey topic. These papers are then ranked by semantic similarity based on their abstracts and the given topic. To ensure temporal consistency, we restrict the pool to papers published before the survey’s publication date.  We then identify the 50 most frequently co-cited papers from the retrieved set and add them to the original candidate pool. 

For literature re-ranking, we use the paper’s citation count\footnote{Citation counts are normalized by the number of years since publication to control for citation accumulation bias.}, the sum of the first and last author’s h-index—the last author often being the corresponding or supervising author associated with publication quality \cite{Contributorship2016, Patterns2017}, and the venue’s h-index to represent citation performance, author influence, and venue reputation, respectively. We assign $\gamma = 0.5$ to citation performance, $\beta = 0.3$ to venue reputation, and $\alpha = 0.2$ to author influence. Based on the final weighted scores, we rank the articles and select a final subset of references that matches the reference count of the corresponding human-written survey for evaluation. The weights were chosen based on the author's intuition and preliminary analysis to reflect the relative importance of citation performance, venue reputation, and author influence. While not exhaustive, these values offer a practical starting point for evaluation. Importantly, our framework is modular and supports alternative weight configurations based on downstream needs. 

For structural consistency evaluation, we removed non-content sections such as \textit{``funding''}, \textit{``acknowledgements''}, \textit{``author contributions''}, \textit{``competing interests''}, and \textit{``supplementary material''}.

\section{Evaluation Metrics}
\label{sec:metrics}
\subsection{Metric Formulations}

\textbf{Citation Quality:} We compute the precision, recall, and F1 score of LLM-generated or RAG-retrieved candidate references with human-selected references as follows:
\begin{gather*}
\text{Precision}_{\text{cite}}
= \frac{R_L \cap R_H}{R_L}\\
\text{Recall}_{\text{cite}}
= \frac{R_L \cap R_H}{R_H}\\
\text{F1}_{\text{cite}}
= 2 \times \frac{\text{Precision}_{\text{cite}} \times \text{Recall}_{\text{cite}}}
{\text{Precision}_{\text{cite}} + \text{Recall}_{\text{cite}}}
\end{gather*}
Here, \(R_L\) and \(R_H\) denote the sets of references generated or retrieved by the LLM and those selected by humans, respectively, and \(\cap\) denotes set intersection. A reference is considered a match if its textual similarity exceeds 0.95, as determined from our preliminary experiments.

To evaluate the accuracy of LLM-generated references, we perform title searches to check whether the generated reference yields an exact match with an existing publication in S2ORC databases.

\textbf{Content Quality:} 
To measure Key Point Recall (KPR) \cite{qi2024long2rag} for generated surveys, we first instruct the LLMs to extract key points from the human-written survey. We then verify whether each extracted key point is captured in the corresponding LLM-generated survey using a question-answering (QA) approach.  The KPR is defined  as follows:
\[
KPR(H_i, G) = 
\begin{cases}
1 & \text{if } H_i \text{ is present in } G, \\
0 & \text{otherwise}.
\end{cases}
\]
where \(H_i\) is the \(i\)-th key point extracted in the human-written survey.  \(G\) is the LLM-generated survey. A higher KPR score indicates that the LLM-generated survey covers more key points from the human-written ones.

\begin{table*}[ht]
\centering
\renewcommand{\arraystretch}{1}
\setlength{\tabcolsep}{10pt} 
\small
\begin{tabular}{llcccc}
\Xhline{1pt}
\rowcolor{gray!15}
\textbf{Task} & \textbf{Comparison}
  & \makecell{\textbf{Topic} \\ \textbf{Relevance}}
  & \makecell{\textbf{Information} \\ \textbf{Coverage}}
  & \makecell{\textbf{Critical} \\ \textbf{Analysis}}
  & \makecell{\textbf{Overall} \\ \textbf{Rating}} \\
\hline\hline

\multirow[c]{2}{*}{Task \hyperref[task1]{{1}}}
        & Comparable                              & 33.3\%    & 33.3\%          & 26.7\%  & 20.0\%  \\
        & LLM-Generated > Human-written           & 20.0\%    & 26.7\%          & 26.7\%  & 13.3\%  \\
\midrule
\multirow[c]{2}{*}{Task \hyperref[task2]{{2}}}
        & Comparable                              & 33.3\%    & 46.7\%          & 40.0\%  & 26.7\%  \\
        & LLM-Generated > Human-written           & 33.3\%    & 20.0\%          & 20.0\%  & 13.3\%  \\
\midrule
\multirow[c]{2}{*}{Task \hyperref[task3]{{3}}}
        & Comparable                              & 40.0\%    & 53.3\%          & 46.7\%  & 26.7\%  \\
        & LLM-Generated > Human-written           & 26.7\%    & 20.0\%          & 20.0\%  & 20.0\%  \\
\bottomrule
\end{tabular}
\caption{Human evaluation results across tasks. Each task includes five surveys from the Computer Science domain, all generated using Claude-3.7-Sonnet. For Task 2, the surveys were generated from the QUAL-SG pipeline.}
\label{tab:human}
\end{table*}

\textbf{Structural Consistency:} 
In structural consistency evaluation, for the structural overlap, we set the semantic similarity threshold to 0.8, as our preliminary experiments showed it to be optimal for identifying valid matches.  The calculation formula is defined as follows:
\[
M(S_H^i, S_G^j) = 
\begin{cases}
1 & \text{if } S_H^i \text{ and } S_G^j \text{ are matching}, \\
0 & \text{otherwise}.
\end{cases}
\]
where  \( S_H^i \)represents the \(i\)-th section from the human-written survey. \( S_G^j \) represents the \(j\)-th section from the LLM-generated survey.
We define the structural consistency between the two as follows:
\[
S_{\text{struct}} = \frac{2 \times (S_H \cap S_G)}{S_H + S_G}
\]
where \( S_H \), \( S_G \) represent the number of sections in the human-written and LLM-generated surveys, respectively, \( S_H \cap S_G \) is the number of matching sections between the two. 

We then use LLM-as-judge to score structural relevance between the LLM-generated survey and the corresponding human-written ones as follows:
\begin{gather*}
\text{Relevance}_{\text{LLM}} = \frac{1}{|S|} \sum_{s \in S} \mathbb{I}_{\text{relevant}}(s, H)
\label{eq:relevance_llm}
\end{gather*}

\subsection{Comparison with Existing Metrics} The \underline{\textit{two main differences}} in our evaluation design compared to prior work \citep{wang2024autosurvey,liang2025surveyx,tang2024llms} are as follows:  

First, in citation quality evaluation, previous studies primarily assess recall—i.e., how many human-selected citations are recovered by the LLM-generated or RAG-based retrieval. However, we argue that measuring recall alone may overestimate model performance. For example, an LLM might generate 8 out of 10 citations from a human-written survey (80\% recall), but also includes over 20 additional irrelevant references; the overall citation reliability is significantly compromised. Therefore, we further introduce citation precision to provide a more balanced assessment of citation quality.

Second, in structural evaluation, prior work mainly examines how well the content aligns with the research topic. In contrast, we directly compare the LLM-generated outline with the survey structure of human-written surveys. This is motivated by the fact that human-written outlines, which are carefully designed and peer-reviewed, better reflect the scope and logic of the survey. Taking human-written surveys as a gold standard, this finer-grained structural evaluation enables a more precise assessment to identify which types of sections are well-covered, missing, or over-generated by the LLMs, helping reveal the specific aspects where LLMs still fall short.

\begin{figure}[htbp]
  \centering
  \includegraphics[width=\linewidth]{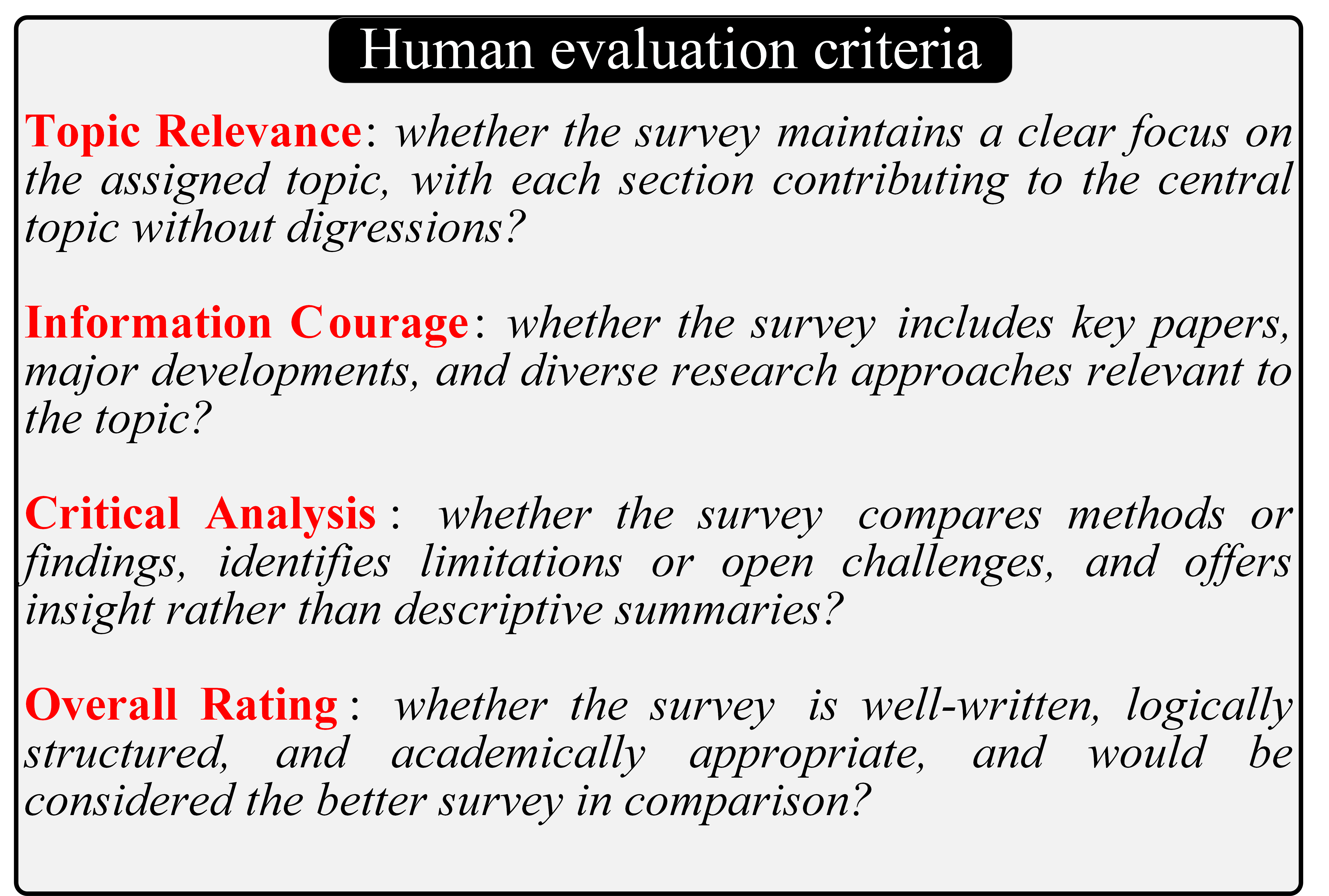}
  \caption{Human evaluation criteria}
  \label{fig:criteria}
\end{figure}
\vspace{-2em}  

\section{Human Evaluation Protocol}
\label{sec:human evaluation}
Given that literature survey evaluation requires specific domain expertise and is time-consuming, we randomly select 5 surveys from the computer science domain for each task, resulting in a total of 15 surveys for human evaluation. Each LLM-generated survey is paired with its corresponding human-written survey to form an evaluation pair. We then invite three second-year PhD students with a background in computer science as annotators, each of whom has published at least one peer-reviewed paper, to compare the LLM-generated and human-written surveys from the following four aspects: \textbf{\textit{topic relevance}}, \textbf{\textit{information coverage}}, \textbf{\textit{critical analysis}}, and \textbf{\textit{overall rating}}. For each pair, annotators are asked to compare the two surveys and judge which one is better, comparable, or worse. To mitigate potential bias, all identifying information was removed, and annotators \textbf{\underline{were not informed}} whether the surveys were LLM-generated or human-written. The human evaluation criteria are illustrated in Figure~\ref{fig:criteria}, and the corresponding evaluation results are summarized in Table~\ref{tab:human} of Section~\ref {sec:human_results}.

\section{Additional Results} 
\subsection{Ablation Study}

\begin{table*}[!htbp]
\centering
\renewcommand{\arraystretch}{1} 
\setlength{\tabcolsep}{8pt}

\begin{tabular}{lccc}
\Xhline{1pt}
\rowcolor{gray!15}
\textbf{Ablation Setting} & \textbf{P ↑} & \textbf{R ↑} & \textbf{F1 ↑} \\
\hline
\hline

\rowcolor{blue!12}
QUAL-SG            & \textbf{15.87} & \textbf{17.71} & \textbf{16.73}   \\
\quad \textit{w/o} co-cited expansion     & 10.07 {\footnotesize\textcolor{red}{(↓5.80)}} & 11.52 {\footnotesize\textcolor{red}{(↓6.19)}} & 10.75 {\footnotesize\textcolor{red}{(↓5.98)}}  \\
\quad \textit{w/o} topical relevance     & 11.54 {\footnotesize\textcolor{red} {(↓4.33)}} & 13.15 {\footnotesize\textcolor{red}{(↓4.56)}} & 12.29 {\footnotesize\textcolor{red}{(↓4.44)}}  \\
\quad \textit{w/o} academic impact       & 8.76 {\footnotesize\textcolor{red}{(↓7.11)}}  & 9.28 {\footnotesize\textcolor{red}{(↓8.43)}}  & 9.01 {\footnotesize\textcolor{red}{(↓7.72)}}   \\
\quad \textit{w/o} content diversity     & \underline{13.16 }{\footnotesize\textcolor{red}{(↓2.71)}} & \underline{14.34} {\footnotesize\textcolor{red}{(↓3.37)}} & \underline{13.72} {\footnotesize\textcolor{red}{(↓3.01)}}  \\
\bottomrule
\end{tabular}
\caption{Ablation study of QUAL-SG in the literature retrieval stage. The best results are marked \textbf{bold} and the second-best are \underline{underlined}.}
\label{tab:AblationStudy}
\end{table*}

\begin{table*}[htbp]
  \centering
  \renewcommand\arraystretch{1}
  \resizebox{\textwidth}{!}{%
    \begin{tabular}{%
      >{\centering\arraybackslash}m{1.6cm}  
      >{\centering\arraybackslash}m{1.4cm} |
      >{\centering\arraybackslash}m{1.4cm}
      >{\centering\arraybackslash}m{1.4cm}
      >{\centering\arraybackslash}m{1.4cm}
      >{\centering\arraybackslash}m{1.8cm} |  
      >{\centering\arraybackslash}m{10.2cm}@{}}
      \Xhline{1pt}
      \rowcolor{gray!15}
      \multicolumn{2}{c|}{\textbf{Key References}} 
      & \multicolumn{4}{c|}{\textbf{Models}} 
       & \multirow{2}{*}[0pt]{\centering\makecell{\textbf{} \\ \textbf{}}} \\
      
      \rowcolor{gray!15}
      \makecell{\textbf{S2ORC} \\ \textbf{ID}} 
      & \makecell{\textbf{Citation} \\ \textbf{Count}} 
      & \makecell{\textbf{Human} \\ \textbf{Selected}} 
      & \makecell{\textbf{Fully-} \\ \textbf{LLMGen}} 
      & \makecell{\textbf{Naive-} \\ \textbf{RAG}} 
      & \makecell{\textbf{QUAL-SG} \\ \textbf{(Ours)}}
      & \makecell{\textbf{Generated} \\ \textbf{Context}}\\

      \midrule
      \rowcolor{blue!12}
      \multicolumn{7}{c}{Title: Deep Learning for Computer Vision: A Brief Review \cite{voulodimos2018deep}}
       \\
      \midrule
      57246310 & 61709 & \checkmark & \checkmark & $\times$ & \checkmark* 
        & \textit{…, the availability of large annotated datasets,}
          \textcolor{purple}{\textit{exemplified by} \textbf{[ref]}}
          \textit{, provided…} \\
      10328909 & 60469 & \checkmark & $\times$ & $\times$ & \checkmark* 
        & \textit{…and its variants are commonly employed}
          \textcolor{orange}{\textit{in object detection frameworks Faster R-CNN}
            \textbf{[ref].}} \\
      2930547  & 38960 & \checkmark & \checkmark & $\times$ & \checkmark* 
        & \textit{…achieving unprecedented accuracy}
          \textcolor{blue}{\textit{on the ImageNet Large Scale Visual Recognition}
            \textbf{[ref].}} \\
      2309950  & 16043 & \checkmark & $\times$ & $\times$ & \checkmark*
        & \textit{…and seminal work like}
          \textcolor{red}{
            \textit{Spatial Pyramid Pooling in Deep Convolutional Networks for Visual Recognition}
            \textbf{[ref]}
          } \\
      \midrule
      \rowcolor{blue!12}
      \multicolumn{7}{c}{Title: Role of Microbial Enzymes in the Bioremediation of Pollutants: A Review \cite{Role2011}
      } \\
      \midrule
      83928450 & 812 & \checkmark & \checkmark & $\times$ & \checkmark* 
        & \textit{…among enzymatic methods,}
          \textcolor{blue}{\textit{laccases stand out for phenol degradation and lignin transformation}
          \textbf{[ref]}} \textit{, making them valuable in...} \\
      1624267  & 519 & \checkmark & $\times$ & $\times$ & \checkmark* 
        & \textit{…with recent advances in}
          \textcolor{orange}{\textit{enzyme engineering and DNA shuffling}
          \textbf{[ref]}} \textit{, enhancing...} \\
      84754528 & 119 & \checkmark & $\times$ & \checkmark & \checkmark 
        & \textit{…White rot fungi can degrade chlorinated phenolics}
          \textcolor{red}{\textit{via enzyme systems for paper industry cleanup}
          \textbf{[ref].}} \\
      \bottomrule
    \end{tabular}%
  }
  \caption{Case study with two surveys from our \textit{SurveyGen}.
    The S2ORC ID refers to the article's ID in the S2ORC corpus.
    An * indicates that the paper was not retrieved via semantic similarity but
    was identified as a highly co-cited reference and therefore included
    in the candidate pool.}
  \label{tab:casestudy}
\end{table*}

We present an ablation study of the QUAL-SG framework by individually removing each of the four key components to assess their contributions: \textbf{(1)} co-cited paper expansion; \textbf{(2)} relevance-based ranking; \textbf{(3)} academic impact-based ranking; and \textbf{(4)} content diversity-based ranking. 

As shown in Table \ref{tab:AblationStudy}, the performance of QUAL-SG declines across all ablation settings. Removing academic impact-based ranking (-7.72\%) and co-cited paper expansion (-5.98\%) caused the most significant drops. This highlights the importance of expanding candidate pools via citation analysis and identifying high-impact research for reference selection. Furthermore, topical relevance and content diversity were also shown to contribute positively.

\subsection{Case Analysis}
We conduct a case analysis using two surveys from the Computer Science and Biology domains. As shown in Table \ref{tab:casestudy}, both Fully-LLMGen and Naive-RAG failed to identify several crucial, human-selected references. Notably, Naive-RAG retrieves only one valid reference (out of seven) due to weak semantic similarity between reference abstracts and the topic; however, these papers are frequently cited by other works, indicating their academic influence despite low semantic similarity. QUAL-SG succeeds in recovering all key papers through two core strategies: first, expanding the candidate pool via co-citation analysis, which allows the inclusion of semantically distant yet influential works; and second, ranking candidates by quality to identify the most impactful studies and better highlight their contributions in the generated survey.

\clearpage
\clearpage
\section{Topic Examples for Survey Generation}
\label{sec:testdata} 
\begin{table}[!htb]
  \centering
  \renewcommand\arraystretch{1}
  \begin{tabular}{lcl}
    \Xhline{0.5pt}
    \rowcolor{gray!15}
    \textbf{S2ORC ID} & \textbf{Topic} & \textbf{Citation} \\
    \midrule
    \rowcolor{blue!12}
    \multicolumn{3}{c}{\textbf{Biology}} \\
    13599358 & Microbial Enzymes in Pollutant Bioremediation & 628 \\
    11116464 & Lactic Acid Bacteria and Bacteriocins & 457 \\
    6209474  & Mathematical Models of Malaria & 361 \\
    5068313  & Effects of Deoxynivalenol and Type B Trichothecenes on the Intestine & 284 \\
    19692413 & Perio-Pathogenic Bacteria in Oral Carcinogenesis & 187 \\
    15915856 & Monosodium Glutamate Toxic Effects and Implications for Human Intake & 140 \\
    17610865 & Marine N-3 Fatty Acids and Type 2 Diabetes Risk & 129 \\
    39756789 & Neonicotinoid Insecticides and Developmental Neurotoxicity & 109 \\
    220843996 & Detection of Human Intestinal Protozoan Parasites in Vegetables and Fruits & 89 \\
    1100406  & Cyanobacterial Natural Products: Structure, Properties, and Applications & 87 \\
    \midrule
    \rowcolor{blue!12}
    \multicolumn{3}{c}{\textbf{Medicine}} \\
    17136958  & Dietary Sugars and Body Weight in Randomised Controlled Trials & 1583 \\
    52095775  & Flavonoids and Phenolic Compounds from Medicinal Plants & 1408 \\
    263077223 & Patient Engagement in Research & 1135 \\
    17464731  & Delirium Outcomes in Critically Ill Patients & 793 \\
    212709676 & Traditional Chinese Medicine in Treating SARS-CoV-2 Infections & 769 \\
    4893818   & Education and Dementia in the Context of the Cognitive Reserve Hypothesis & 740 \\
    51958985  & ADHD Medications: Efficacy and Tolerability & 705 \\
    13897386  & Short Term Air Pollution Exposure and Stroke & 705 \\
    219607020 & Amyotrophic Lateral Sclerosis: Clinical Perspectives & 657 \\
    6017773   & Maternal Smoking During Pregnancy and Associated Birth Defects & 626 \\
    \midrule
    \rowcolor{blue!12}
    \multicolumn{3}{c}{\textbf{Psychology}} \\
    1845793   & Neuroimaging Studies of Internet and Gaming Addiction & 377 \\
    52293261  & Technology-Delivered Interventions for Youth Depression and Anxiety & 212 \\
    18781074  & Music Therapy and Cognitive Function in Alzheimer's Disease & 174 \\
    20918937  & Gender Dysphoria and Autism Spectrum Disorder & 173 \\
    4033267   & Figurative Language Comprehension in Autism Spectrum Disorder & 172 \\
    2802244   & Propranolol in Anxiety Disorder Treatment & 170 \\
    4152509   & ADHD Prevalence in Chinese Children and Adolescents & 161 \\
    54447675  & Influence of Role Models on Gender and Careers & 156 \\
    3916988   & Motivation in Health Education and Self-Determination Theory & 152 \\
    12043081  & Fundamental Criteria for Eating Disorder Recovery & 141 \\
    \midrule
    \rowcolor{blue!12}
    \multicolumn{3}{c}{\textbf{Computer Science}} \\
    10137425  & Multimodal Machine Learning Taxonomy & 2737 \\
    3557281   & Deep Learning Applications in Computer Vision & 2709 \\
    218474694 & IoT Sensing with RFID and Wireless Sensor Networks & 269 \\
    235410640 & Deep Multimodal Learning for Computer Vision & 266 \\
    258180322 & Fairness and Bias in Artificial Intelligence & 216 \\
    232300174 & Facial Micro-Expression Analysis & 180 \\
    231802191 & Synthetic-CT Generation in Radiotherapy and PET & 131 \\
    208035941 & Detecting Sleep Apnea Using Deep Learning & 115 \\
    186207561 & Head-Mounted Eye Gaze Tracking Devices & 99 \\
    150036628 & Electrical Impedance Tomography and AI Applications & 90 \\
    \bottomrule
  \end{tabular}
  \label{tab:topic-examples}
\end{table}
\clearpage

\section{SurveyGen Data Format}\label{sec:dataformat}  

\begin{figure}[htbp]
  \centering
  \includegraphics[trim=1 1 1 1, clip, scale=0.7]{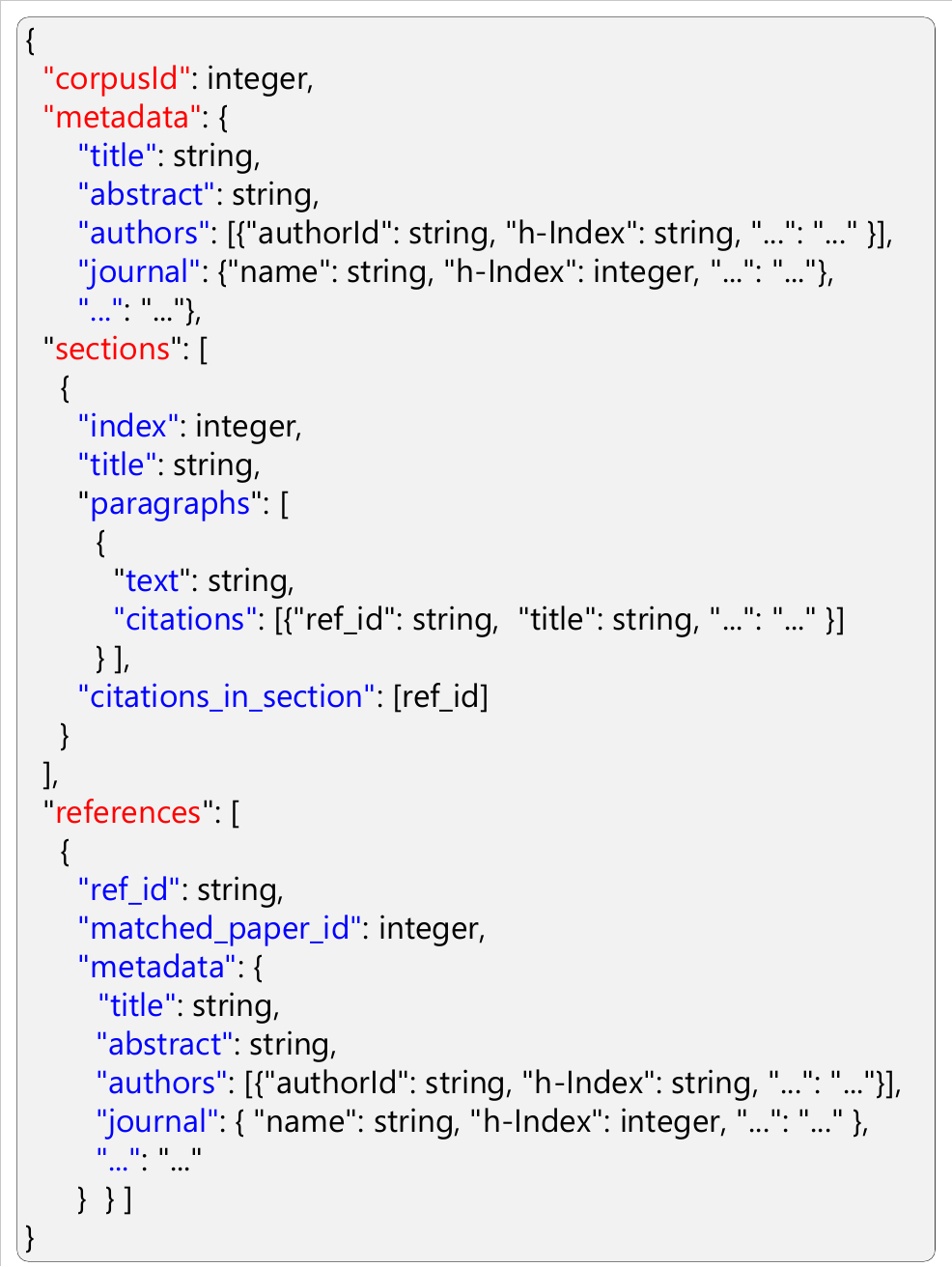}
  \label{fig:data-format}
\end{figure}

\clearpage

\section{Prompt used in this study}
\vspace{2em}
\begin{tcolorbox}[
    enhanced,
    colback=gray!10,            
    colframe=black,             
    coltitle=white,             
    colbacktitle=black,         
    title=Prompt for Survey-type Article Classification,
    fonttitle=\bfseries\large,  
    width=\textwidth,
    boxrule=0.5pt,
    arc=2pt,
    outer arc=2pt,
    left=5pt,
    right=5pt,
    top=5pt,
    bottom=5pt
]

You are an academic expert in scientific literature classification. Your task is to determine whether the following paper is a survey-type article, based on its title and abstract below:

\vspace{0.8em}
\textit{Title: \{title\}}\\
\textit{Abstract: \{abstract\}}

\vspace{0.8em}
Please make your judgment based on the following three criteria:

1. The abstract explicitly declares a survey intent (e.g., phrases like \textit{``conducts a survey'', ``provides an overview'', ``this survey...''}, etc.).

2. The focus of the paper is on reviewing or summarizing existing work, rather than proposing new methodologies or reporting novel experimental results.

3. The abstract provides a forward-looking perspective on the field by synthesizing the reviewed literature to identify key trends, highlight significant open challenges or gaps, and suggest promising directions for future investigation.

\vspace{0.8em}
Your output should be only one word: \textbf{``TRUE'' } if the paper qualifies as a survey-type article, or \textbf{``FALSE''} if it does not. Do not include any additional commentary, explanation, or formatting instructions.

\end{tcolorbox}

\vspace{5em}
\begin{tcolorbox}[
    enhanced,
    colback=gray!10,            
    colframe=black,             
    coltitle=white,             
    colbacktitle=black,         
    title=Prompt for Topic Relevance Evaluation,
    fonttitle=\bfseries\large,  
    width=\textwidth,
    boxrule=0.5pt,
    arc=2pt,
    outer arc=2pt,
    left=5pt,
    right=5pt,
    top=5pt,
    bottom=5pt
]

You are an academic expert helping to write a survey on the topic: "{\textbf{\{TOPIC\}}}".

You will be provided with the title and abstract of a research paper. Your task is to rating the paper's relevance to the survey topic  based on the following criteria:

\vspace{0.8em}
\textbf{Score-1 (poor)}: The paper is unrelated to the topic or only mentions it in passing, with no meaningful contribution.

\textbf{Score-2 (low)}: The paper is loosely connected or provides general background, but not focused on the topic.

\textbf{Score-3 (moderate)}: The paper discusses a specific sub-aspect of the topic; somewhat useful, but not central.

\textbf{Score-4 (high)}: The paper substantially addresses key elements of the topic; would likely be cited in the survey.

\textbf{Score-5 (very high)}: The paper is entirely focused on the topic, offering essential insights; likely foundational to the survey.

\vspace{0.8em}
Your output should only include a single score from 1 to 5. Do not provide any explanation or additional text.

\vspace{0.8em}
Title:{title of the paper}
Abstract:{abstract of the paper}

\vspace{0.8em}
Topical relevance score:
\end{tcolorbox}

\clearpage

\begin{tcolorbox}[
    enhanced,
    colback=gray!10,        
    colframe=black,         
    coltitle=white,         
    colbacktitle=black,     
    title=Prompt for Outline Generation (Task 1),
    fonttitle=\bfseries,    
    width=\textwidth,       
    boxrule=0.5pt,
    arc=2pt,
    outer arc=2pt,
    left=5pt,
    right=5pt,
    top=5pt,
    bottom=5pt
]

You are an academic expert in the field of "{\textbf{\{TOPIC\}}}" with deep expertise in survey writing. Your task is to generate a well-structured outline for a survey on this topic.

\vspace{0.8em}
\textbf{Please follow the instructions below:}
\vspace{0.6em}

\textbf{Step 1:} Based on the given topic, identify 3 to 7 major thematic sections that define the overall scope and objectives of the survey. For each section, provide an academically styled title, along with a brief description summarizing its focus and relevance.

\vspace{0.8em}
\textbf{Step 2:} For the major thematic sections, list several subsections representing more specific research areas, concepts, or points to be covered. Subsections should be conceptually related to their parent section and serve to further structure the survey.

\vspace{0.8em}
\textbf{Step 3:} Your output should be in JSON format and must include the survey title, a structured outline with section titles, descriptions, and subsections. Ensure the structure is logically coherent, well-aligned with the topic, and suitable for developing a full-length academic survey.

\vspace{0.8em}
The output example should follow the format below:

\begin{verbatim}
{
  "title": "TITLE OF THE SURVEY",
  "outline": [
    {
      "section_title": "SECTION TITLE 1",
      "description": "A brief description of this section.",
      "subsections": {
        "subsection title 1": "content",
        "...": "...",
        "subsection title n": "content"
      }
    },
    {
      "section_title": "SECTION TITLE 2",
      "description": "A brief description of this section.",
      "subsections": {
        "subsection title 1": "content",
        "...": "...",
        "subsection title n": "content"
      }
    }
  ]
}
\end{verbatim}

Now, based on the given topic \{Topic\}, please generate the outline by following the steps above. Do not include any additional commentary, explanation, or formatting instructions; only return the structured JSON output as specified.

\end{tcolorbox}

\clearpage

\begin{tcolorbox}[
    enhanced,
    colback=gray!10,        
    colframe=black,         
    coltitle=white,         
    colbacktitle=black,     
    title=Prompt for Section Content Generation (Task 1),
    fonttitle=\bfseries,    
    width=\textwidth,       
    boxrule=0.5pt,
    arc=2pt,
    outer arc=2pt,
    left=5pt,
    right=5pt,
    top=5pt,
    bottom=5pt
]

You are an academic expert in the field of "{\textbf{\{TOPIC\}}}"  with deep expertise in survey writing. Your task is to write a subsection of a survey, based on the following information:

\vspace{0.8em}
The overall structure of the survey is as follows:  
\begin{itemize}
    \item \{\{outline\}\}
\end{itemize}

You are now asked to write the following subsection:  
\begin{itemize}
    \item Subsection: \{\{subsection\_title\}\}
\end{itemize}

\vspace{0.6em}

\textbf{Instructions for generating subsection content:}

1. Ensure the content directly addresses the specific topic of the subsection, and the generated content should be a minimum of 300 words.

\vspace{0.3em}  
2. Ensure that all claims are fully supported by relevant academic literature. Cite each reference using in-text citations in the format \texttt{ref [1]}, \texttt{ref [2]}, etc. If a source is cited multiple times, use only the reference number assigned to its first occurrence.

\vspace{0.3em}     
3. Maintain alignment with the parent section and overall survey topic, ensuring thematic and conceptual consistency.

\vspace{0.3em}     
4. Use a formal academic tone, with logically structured arguments and scholarly language.

\vspace{0.3em}    
5. Your output should strictly be a JSON object with the following two fields.

\begin{verbatim}
{
    "content": "the content of the subsection",
    "references": [
    {
        "refNo": "reference number",
        "authors": "full author list",
        "year": "year of publication",
        "title": "title of the paper",
        "venue": "publication source",
        "doi": "DOI"
    }
    ]
}
\end{verbatim}

Now, please generate the content for the given subsection: \{subsection\_title\}. Please ensure the output is formatted according to the requirements mentioned above. Do not include any explanation, commentary, or preamble in your response.

\end{tcolorbox}

\clearpage

\begin{tcolorbox}[
    enhanced,
    colback=gray!10,        
    colframe=black,         
    coltitle=white,         
    colbacktitle=black,     
    title=Prompt for Outline Generation (Task 2),
    fonttitle=\bfseries,    
    width=\textwidth,       
    boxrule=0.5pt,
    arc=2pt,
    outer arc=2pt,
    left=5pt,
    right=5pt,
    top=5pt,
    bottom=5pt
]

You are an academic expert in the field of "{\textbf{\{TOPIC\}}}" with deep expertise in survey writing. Your task is to generate a well-structured outline for a survey on this topic.

\vspace{0.8em}
The following highly relevant papers, including their titles and abstracts, are provided for reference:

\begin{itemize}
    \item \{\{references list\}\}
\end{itemize}

\vspace{0.6em}
Please follow the instructions below:

\vspace{0.6em}
\textbf{Step 1:} Based on the given topic, identify 3 to 7 major thematic sections that define the overall scope and objectives of the survey. For each section, provide an academically styled title, along with a brief description summarizing its focus and relevance.

\vspace{0.8em}
\textbf{Step 2:} For the major thematic sections, list several subsections representing more specific research areas, concepts, or points to be covered. Subsections should be conceptually related to their parent section and serve to
further structure the survey.

\vspace{0.8em}
\textbf{Step 3:} Your output should be in JSON format and must include the survey title, a structured outline with section titles, descriptions, and subsections. Ensure the structure is logically coherent, well-aligned with the topic, and suitable for developing a full-length academic survey.

\vspace{0.8em}
The output example should follow the format below:

\begin{verbatim}
{
  "title": "TITLE OF THE SURVEY",
  "outline": [
    {
      "section_title": "SECTION TITLE 1",
      "description": "A brief description of this section.",
      "subsections": {
        "subsection title 1": "content",
        "...": "...",
        "subsection title n": "content"
      }
    },
    {
      "section_title": "SECTION TITLE 2",
      "description": "A brief description of this section.",
      "subsections": {
        "subsection title 1": "content",
        "...": "...",
        "subsection title n": "content"
      }
    }
  ]
}
\end{verbatim}
\vspace{0.6em} 
Now, based on the given topic \{Topic\}, please generate the outline by following the steps above. Do not include any additional commentary, explanation, or formatting instructions; only return the structured JSON output as specified.

\end{tcolorbox}

\clearpage

\begin{tcolorbox}[
    enhanced,
    colback=gray!10,        
    colframe=black,         
    coltitle=white,         
    colbacktitle=black,     
    title=Prompt for Section Content Generation (Task 2),
    fonttitle=\bfseries,    
    width=\textwidth,       
    boxrule=0.5pt,
    arc=2pt,
    outer arc=2pt,
    left=5pt,
    right=5pt,
    top=5pt,
    bottom=5pt
]

You are an academic expert in the field of "{\textbf{\{TOPIC\}}}"  with deep expertise in survey writing. Your task is to write a subsection of a survey, based on the following information:

\vspace{0.8em}
The overall structure of the survey is as follows:  
\begin{itemize}
    \item \{\{outline\}\}
\end{itemize}

You are now asked to write the following subsection:  
\begin{itemize}
    \item Subsection: \{\{subsection\_title\}\}
\end{itemize}

The following highly relevant papers, including their titles and abstracts, are provided for reference:
\begin{itemize}
    \item Subsection: \{\{references\_list\}\}
\end{itemize}

\vspace{0.6em}

\textbf{Instructions for generating subsection content:}

\vspace{0.3em}  
1. Carefully analyze the provided references and identify those highly relevant to the subsection topic as the basis for your generation. 

\vspace{0.3em}  
2. Ensure the content directly addresses the specific topic of the subsection, and the generated content should be a minimum of 300 words.

\vspace{0.3em}  
3. Ensure that all claims are fully supported by relevant academic literature. Cite each reference using in-text citations in the format \texttt{ref [1]}, \texttt{ref [2]}, etc. If a source is cited multiple times, use only the reference number assigned to its first occurrence.

\vspace{0.3em}     
4. Maintain alignment with the parent section and overall survey topic, ensuring thematic and conceptual consistency.

\vspace{0.3em}     
5. Use a formal academic tone, with logically structured arguments and scholarly language.

\vspace{0.3em}    
6. Your output should strictly be a JSON object with the following two fields.

\begin{verbatim}
{
    "content": "the content of the subsection",
    "references": [
    {
        "refNo": "reference number",
        "title": "title of the paper",
    }
    ]
}
\end{verbatim}

\vspace{0.6em}   
Now, please generate the content for the given subsection: \{subsection\_title\}. Please ensure the output is formatted according to the requirements mentioned above. Do not include any explanation, commentary, or preamble in your response.

\end{tcolorbox}

\clearpage
\begin{tcolorbox}[
    enhanced,
    colback=gray!10,        
    colframe=black,         
    coltitle=white,         
    colbacktitle=black,     
    title=Prompt for Section Content Generation (Task 3),
    fonttitle=\bfseries,    
    width=\textwidth,       
    boxrule=0.5pt,
    arc=2pt,
    outer arc=2pt,
    left=5pt,
    right=5pt,
    top=5pt,
    bottom=5pt
]

You are an academic expert in the field of "{\textbf{\{TOPIC\}}}"  with deep expertise in survey writing. Your task is to write a subsection of a survey, based on the following information:

\vspace{0.3em}
The overall structure of the survey is as follows:  
\begin{itemize}[nosep]
    \item \{\{outline\}\}
\end{itemize}

You are now asked to write the following subsection:  
\begin{itemize}[nosep]
    \item Subsection: \{\{subsection\_title\}\}
\end{itemize}

The following papers are provided for this subsection and should be summarized accordingly:
\begin{itemize}[nosep]
    \item \{\{references\_list\}\}
\end{itemize}

\vspace{0.3em}
\textbf{Instructions for generating subsection content:}

\vspace{0.2em}
1. You must generate the content based on the provided references. Do not incorporate or cite any external references.

\vspace{0.2em} 
2. Ensure the content directly addresses the specific topic of the subsection, and the generated content should be a minimum of 300 words

\vspace{0.2em} 
3. Ensure that all claims are fully supported by relevant academic literature. Cite each reference using in-text citations in the format \texttt{ref [1]}, \texttt{ref [2]}, etc. If a source is cited multiple times, use only the reference number assigned to its first occurrence.

\vspace{0.2em}     
4. Maintain alignment with the parent section and overall survey topic, ensuring thematic and conceptual consistency.

\vspace{0.2em}     
5. Use a formal academic tone, with logically structured arguments and scholarly language.

\vspace{0.4em}  
Now, please generate the content for the given subsection: \{subsection\_title\}. Please ensure the output is formatted according to the requirements mentioned above. Do not include any explanation, commentary, or preamble in your response.
\end{tcolorbox}

\vspace{0.4em}
\begin{tcolorbox}[
    enhanced,
    colback=gray!10,            
    colframe=black,             
    coltitle=white,             
    colbacktitle=black,         
    title=Prompt for Structural Consistency Evaluation,
    fonttitle=\bfseries\large,  
    width=\textwidth,
    boxrule=0.5pt,
    arc=2pt,
    outer arc=2pt,
    left=5pt,
    right=5pt,
    top=5pt,
    bottom=5pt
]

You are an academic expert tasked with evaluating a draft outline for a literature survey on the topic of "{\textbf{\{TOPIC\}}}".

\vspace{0.4em}   
You will be provided with two outlines:

\textbf{1}. A draft outline as a preliminary version. \\
\quad -- \textbf{LLM-generated Outline}

\textbf{2}. A gold-standard outline created by domain experts. \\
\quad -- \textbf{Human-written Outline}

\vspace{0.2em}  
Your task is to rate the structural consistency between the two outlines following the criteria below:

\vspace{0.4em}  
\textbf{Score 1 (Very Poor)}: The draft outline is largely inconsistent with the gold-standard; major sections are missing or irrelevant.

\textbf{Score 2 (Poor)}: The draft outline only partially overlaps with the gold-standard; some key themes are omitted or misplaced.

\textbf{Score 3 (Moderate)}: The draft includes some of the core topics but lacks structural alignment or completeness.

\textbf{Score 4 (Good)}: The draft mostly aligns with the gold-standard; minor deviations in structure or scope.

\textbf{Score 5 (Excellent)}: The draft closely mirrors the gold standard in both structure and content coverage.

\vspace{0.4em}  
Your output should only include a single score from 1 to 5. Do not provide any explanation or additional text.

\end{tcolorbox}

\end{document}